\documentclass[table]{article}

\usepackage{microtype}
\usepackage{graphicx}
\usepackage{subfigure}
\usepackage{float}
\usepackage{booktabs} % for professional tables

\usepackage{hyperref}

\usepackage{multirow}
\usepackage[accepted]{icml2024}
\usepackage{amsmath}
\usepackage{enumitem}
\usepackage{amssymb}
\usepackage{mathtools}
\usepackage{amsthm}
\definecolor{DarkGreen}{RGB}{17,193,56}
\usepackage{pifont}% http://ctan.org/pkg/pifont
\newcommand{\cmark}{\textcolor{DarkGreen}{\ding{51}}}%
\newcommand{\xmark}{\textcolor{red}{\ding{55}}}%

\usepackage[capitalize,noabbrev]{cleveref}

\theoremstyle{plain}

\theoremstyle{definition}

\theoremstyle{remark}

\usepackage[textsize=tiny]{todonotes}

\icmltitlerunning{MagicPose: Realistic Human Pose and Facial Expression Retargeting with Identity-aware Diffusion}

\begin{document}

\twocolumn[
\icmltitle{MagicPose: Realistic Human Poses and Facial Expressions Retargeting with Identity-aware Diffusion}




\icmlsetsymbol{equal}{*}

\begin{icmlauthorlist}
\icmlauthor{Di Chang}{usc,tt}
\icmlauthor{Yichun Shi}{tt}
\icmlauthor{Quankai Gao}{usc}
\icmlauthor{Jessica Fu}{usc}
\icmlauthor{Hongyi Xu }{tt}\\
\icmlauthor{Guoxian Song}{tt}
\icmlauthor{Qing Yan}{tt}
\icmlauthor{Yizhe Zhu }{tt}
\icmlauthor{Xiao Yang }{tt}
\icmlauthor{Mohammad Soleymani}{usc}

% % \url{https://boese0601.github.io/magicdance/} \\
% {\tt\small dichang@usc.edu}\\


\end{icmlauthorlist}
\icmlaffiliation{usc}{University of Southern California}
\icmlaffiliation{tt}{ByteDance Inc}
% \icmlcorrespondingauthor{Mohammad Soleymani}{soleymani@ict.usc.edu}
\icmlcorrespondingauthor{Di Chang}{dichang@usc.edu}

\icmlkeywords{Diffusion Model, Human Pose Retargeting}

\vskip 0.3in

]



\printAffiliationsAndNotice{} 
% leave blank if no need to mention equal contribution

%\printAffiliationsAndNotice{\icmlEqualContribution} 
% otherwise use the standard text.

% \begin{center}
%     \centering
%     \captionsetup{type=figure}
%     \includegraphics[width=\linewidth]{Figures/Exps/CVPR_Teaser_cropped.pdf}
%     \captionof{figure}{We propose MagicPose, a novel and effective approach to provide realistic human video generation enabling vivid motion and facial expression transfer, and consistent in-the-wild zero-shot generation without any fine-tuning. MagicPose can precisely generate appearance-consistent results, while the original text-to-image model (e.g., Stable Diffusion and ControlNet) can hardly maintain the subject identity information accurately. Furthermore, our proposed modules can be treated as an extension/plug-in to the original text-to-image model without modifying its pre-trained weight. }
%     \label{teaser}
% \end{center}%
% }]

\begin{abstract}
In this work, we propose MagicPose, a diffusion-based model for 2D human pose and facial expression retargeting. Specifically, given a reference image, we aim to generate a person's new images by controlling the poses and facial expressions while keeping the identity unchanged. To this end, we propose a two-stage training strategy to disentangle human motions and appearance (e.g., facial expressions, skin tone and dressing), consisting of (1) the pre-training of an appearance-control block and (2) learning appearance-disentangled pose control. Our novel design enables robust appearance control over generated human images, including body, facial attributes, and even background. By leveraging the prior knowledge of image diffusion models, MagicPose generalizes well to unseen human identities and complex poses without the need for additional fine-tuning. Moreover, the proposed model is easy to use and can be considered as a plug-in module/extension to Stable Diffusion. The code is available at \url{https://github.com/Boese0601/MagicDance}.
\end{abstract}
   
\section{Introduction}\label{sec:intro}
\begin{figure}[t!]
    \centering
    \vspace{-7pt}
    \includegraphics[width=\linewidth]{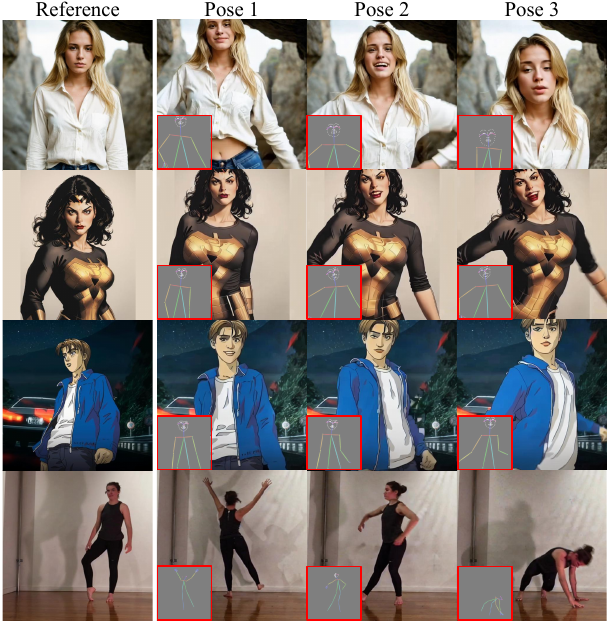}
    \vspace{-20pt}
    \caption{MagicPose can provide zero-shot and realistic human poses and facial expressions retargeting for human images of different styles and poses. A shared model is used here for
    in-the-wild generalization without any fine-tuning on target domains. Our proposed modules can be treated as an extension/plug-in to the original text-to-image model without modifying its pre-trained weight. }
    \label{teaser}
    \vspace{-15pt}
\end{figure}

%\YS{say something else first. Don't start with Image style transfer...}

%\MS {Start with saying what you are solving and write a motivation of why someone should care about this topic. Why is this significant?
%then position it within the context of what has been done in the past (a short, high-level review of the relevant work and their shortcomings that you are solving) 
%how is this method better? What does it do better? - summarize the main points of the work and add major contributions}

Human motion transfer is a challenging task in computer vision. This problem involves retargeting body and facial motions, from one source image to a target image. Such methods can be used for image stylization, editing, digital human synthesis, and possibly data generation for training perception models.

% Given a reference image of a human individual, generating a novel video that has motion or pose sequences transferred from a source video of another human individual to the reference identity boils down to motion (or pose) transfer. This is extremely challenging because the problem is not merely an image-to-image style transfer~\cite{gatys2016image,jing2019neural,luan2017deep} or translation~\cite{isola2017image,zhu2017unpaired}, but also requires both temporal and spatial consistency among frames. 
%Traditionally, some previous methods~\cite{zhou2019dance} considered only image-to-image editing or translation~\cite{isola2017image,zhu2017unpaired}, which is an easier setting. 
%Image editing has witnessed remarkable achievements since the very beginning of computer vision research, from filtering, noise reduction, sharpening, and color adjustments using handcrafted operations to more challenging applications with implicit constraints and complex settings.

%Though Image style transfer is a long-standing task in computer vision applications, including pixel-to-pixel image style transfer usually done by transferring the texture patterns from one image to another. Significant progress~\cite{gatys2016image,jing2019neural,luan2017deep} has also been made on static image style transfer specifically. Illumination, tone curves, and other artistic style can be migrated from reference images to input images while the main semantics of the input images remain unchanged.

Traditionally, human motion transfer is achieved by training a task-specific generative model, such as generative adversarial networks (GANs) on specific datasets, e.g.,~\cite{Siarohin_2018_CVPR,Siarohin_2019_PAMI,liu2019liquid,wei2020gac,sun2022human} for body pose and ~\cite{Wu_2020_CVPR,qiao2018geometry,hong2022depth} for facial expressions. Such methods commonly suffer from two issues: (1) they are typically dependent on an image warping module~\cite{Siarohin_2018_CVPR,Siarohin_2019_PAMI} and hence struggle to interpolate the body parts that are invisible in the reference image due to perspective change or self-occlusion, and (2) they can hardly generalize to images that are different from the training data, greatly limiting their application scope.

Recently, diffusion models~\cite{ho2020denoising,song2020score,rombach2021highresolution,zhang2023adding} have exhibited impressive ability on image generation~\cite{bertalmio2000image,yeh2017semantic,lugmayr2022repaint}. By learning from web-scale image datasets, these models present powerful visual priors for different downstream tasks, such as image inpainting~\cite{lugmayr2022repaint,saharia2022palette,jam2021comprehensive}, video generation~\cite{ho2022imagen,wu2023tune,singer2022make}, 3D generation~\cite{poole2022dreamfusion,raj2023dreambooth3d,shi2023mvdream} and even image segmentations~\cite{amit2021segdiff,baranchuk2021label,wolleb2022diffusion}. Thus, such diffusion priors are great candidates for human motion transfer.
% However, diffusion models still encounter significant challenges in achieving temporal consistency for video generation. Such challenges are further emphasized through the difficult task of motion transfer from single image to video sequences involving complex human motions (e.g. dancing) with unseen pose input. 
%CoDeF~\cite{ouyang2023codef} can generate dynamic scene contents including vivid motions, which are mostly visually appealing but not time-intensive.
Two recent studies, DreamPose~\cite{karras2023dreampose} and DisCo~\cite{wang2023disco}, have attempted to adapt diffusion models for human body re-posing. However, we found that they are still limited in either generation quality, identity preservation (as discussed in Section.~\ref{comparison}), or temporal consistency due to the limits in model design and training strategy. Moreover, there is no clear advantage of these methods over GAN-based methods in generalizability. For example, Disco~\cite{wang2023disco} still needs to be fine-tuned to adapt to images of out-of-domain styles.
% However, DreamPose doesn't consider the interactions between the background and foreground poses, while 
 % DisCo considers the above but has compromising quality and temporal consistency. 

\begin{figure*}[t!]
\centering
 \vspace{-7pt}
 \includegraphics[width=\linewidth]{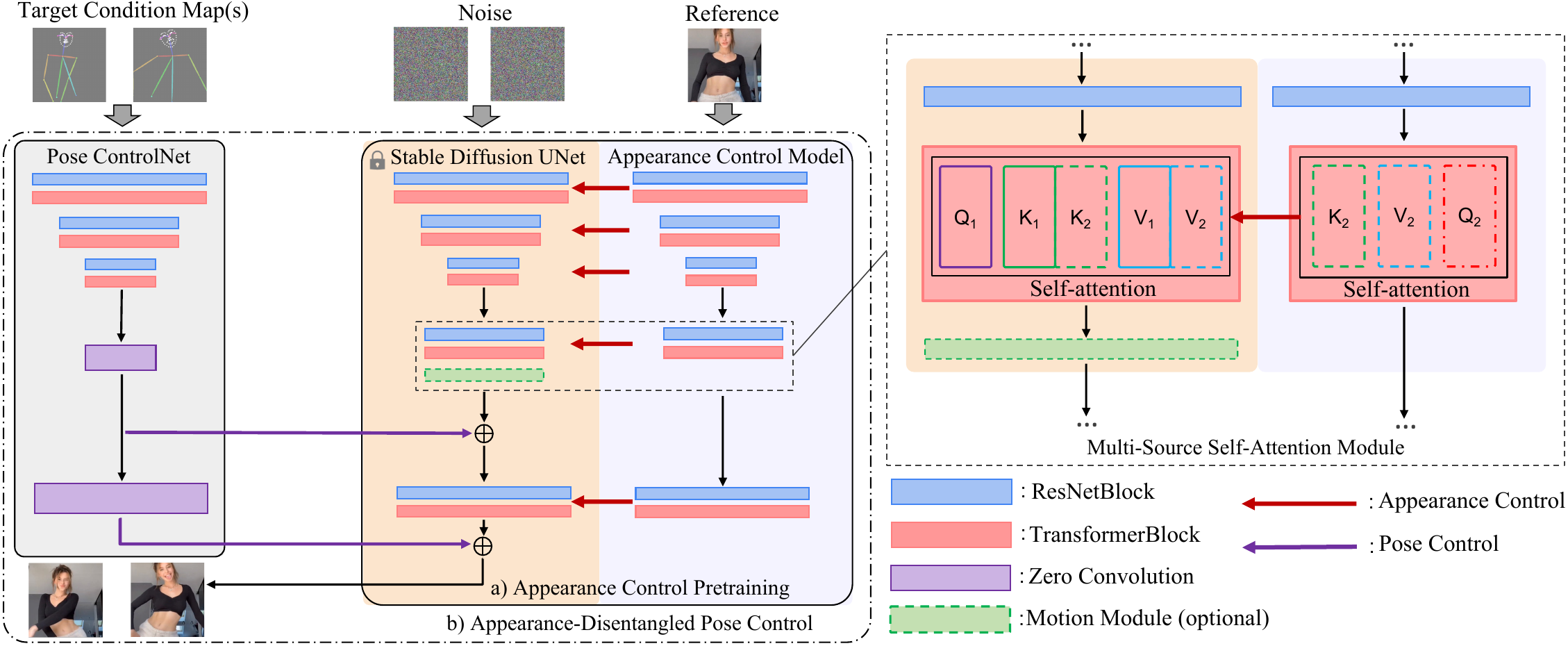}
 \vspace{-20pt}
   \caption{
     Overview of the proposed MagicPose pipeline for controllable human poses and facial expressions retargeting with motions \& facial expressions transfer.  The Appearance Control Model is a copy of the entire Stable-Diffusion UNet, initialized with the same weight. The Stable-Diffusion UNet is frozen throughout the training. During a) Appearance Control Pretraining, we train the appearance control model and its Multi-Source Self-Attention Module. During b) Appearance-disentangled Pose Control, we jointly fine-tune the Appearance Control Model, initialized with weights from a), and the Pose ControlNet. After these steps, an optional motion module can be integrated into the pipeline and fine-tuned for better sequential output generation quality.
   }
   \vspace{-15pt}
    \label{fig:framework}
\end{figure*}

%\MS {This part is not too bad but the major contributions need to be re-written}
In this work, we propose MagicPose to fully exploit the potential of image diffusion priors for human pose retargeting, demonstrating superior visual quality, identity preservation ability, and domain generalizability, as illustrated in Figure.~\ref{teaser}. Our key idea is to decompose the problem into two tasks: (1) identity/appearance control and (2) pose/motion control, which we consider useful capabilities required by image diffusion priors to achieve accurate motion transfer. Correspondingly, as shown in Figure.~\ref{fig:framework}, MagicPose has two sub-modules besides the Stable Diffusion~(SD)~\cite{rombach2021highresolution}: 1) Appearance Control Model that provides appearance guidance from a reference image to the SD via \textbf{Multi-Source Attention Module}, and 
2) Pose ControlNet, which provides pose/expression guidance from a condition image. A multi-stage training strategy is also proposed to effectively learn these sub-modules to disentangle the appearance and pose control.
% Our model training has two stages corresponding to the network design.
% In the first stage, we train our Appearance Control Model on input video sequences focusing on human identity control regardless of spatial varying motions. In the second stage, we train our Pose ControleNet for motion controlling only while fine-tuning the Appearance Control Model as our Appearance-disentangled Pose Control.
Extensive experiments demonstrate the effectiveness of MagicPose which can retain well the key features of the reference identities, including skin tone and clothing, while following the pose skeleton and facial landmark inputs. Moreover, MagicPose can generalize well to unseen identities and motions without any fine-tuning.
The main contributions of this work are as follows:
\begin{itemize}[leftmargin=7pt]\vspace{-0.5em}
    \item An effective method (MagicPose) for human pose and expression retargeting as a plug-in for Stable Diffusion.\vspace{-0.5em} 
    \item Multi-Source Attention Module that offers detailed appearance guidance.\vspace{-0.5em}
    \item A two stage training strategy that enables appearance-pose-disentangled generation.\vspace{-0.5em}
    % \begin{enumerate}
    %     \item Appearance Generalization: MagicPose demonstrates the ability to generate diverse appearances, including those with cartoon-style human identities.
    %     \item Motion Generalization: MagicPose is also capable of creating a wide range of motions, exemplified by its performance on Everybody Dance Now~\cite{chan2019everybody} dataset.
    % \end{enumerate}
    % \item Our model is designed as a plug-in for Stable Diffusion that does not require fine-tuning of the SD parameters and is compatible with existing model weights.\vspace{-0.5em}
    \item Experiment on out-of-domain data demonstrating strong generalizability of our model to diverse image styles and human poses.\vspace{-0.5em}
    \item Comprehensive experiments conducted on TikTok dataset showing model's superior performance in pose retargeting.

\end{itemize}

\section{Related Work}

\label{sec:relatedwork}
\subsection{Human Motion/Expression Transfer} 
Early work in human motion transfer primarily involved manipulation of given image sequence segments to create a desired action~\cite{10.1145/258734.258880,1238420,10.1145/133994.134003}. Subsequent solutions shifted their focus towards generating three-dimensional (3D) representations of human subjects and performing motion transfer within 3D environments~\cite{1335262,10.1145/2010324.1964927}. However, these approaches were characterized by significant time and labor requirements. In contrast, recent advancements leverage deep learning to learn detailed representations of the input~\cite{Tulyakov:2018:MoCoGAN,kim2018deep,chan2019dance}. This shift has facilitated motion transfer with heightened realism and increased automation. Generative Adversarial Networks (GANs) have been a clear deep learning approach to motion transfer tasks~\cite{albahar2021pose,10.1145/258734.258880,1238420}, providing realistic image generation and Conditional GANs adding further conditioning~\cite{mirza2014conditional}. Kim et al.~\cite{kim2018deep} took synthetic renderings, interior face model, and gaze map to transfer head position and facial expression from one human subject to another, presenting the results as detailed portrait videos. MoCoGAN~\cite{Tulyakov:2018:MoCoGAN} also implements unsupervised adversarial training to perform motion and facial expression transfer onto novel subjects. Chan et al.~\cite{chan2019dance} further advanced this approach to full-body human motion synthesis by utilizing a video-to-video approach, taking in 2D video subjects and 2D pose stick figures to produce transferred dance sequences on new human subjects. In the sub-domain of fashion video synthesis, DreamPose~\cite{karras2023dreampose} used SD with human image input and pose sequence input to generate videos featuring human subjects executing pose sequences with intricate fabric motion. DisCo~\cite{wang2023disco}, another SD-based model, contributed to the use-case of human dance generation, enabling controllable human reference, background reference, and pose maps to produce arbitrary compositions that maintain faithfulness and generalizability to unseen subjects.

\subsection{Image/Video Diffusion Models}
Previous research has demonstrated the effectiveness of diffusion probabilistic models~\cite{song2021denoising,song2021scorebased} for image generation ~\cite{ramesh2022hierarchical,saharia2022photorealistic,DBLP:journals/corr/abs-2112-10741}. Latent diffusion models~\cite{ho2020denoising} have further advanced this domain by reducing computational costs by executing the diffusion step in a lower-dimensional latent space rather than pixel space. With customization and specification being important aspects of content generation, the text-to-image approach has gained popularity as a means of achieving controllable image generation, with notable examples such as Imagen~\cite{saharia2022photorealistic} and SD~\cite{rombach2021highresolution}. The introduction of ControlNet~\cite{zhang2023adding} extended the approach to controllable generation by introducing additional conditioning to SD models, enabling input sources such as segmentation maps, pose key points, and more. Additional condition inputs has enabled a higher degree of customization and task-specificity in the generated outputs, providing a contextual foundation for conditional image generation. With the advancement of conditional image generation, there is a natural extension towards the synthesis of dynamic visual content. Blattmann et al.~\cite{blattmann2023videoldm} showed the use-case of latent diffusion models for video generation by integrating a temporal dimension to the latent diffusion model and further fine-tuning the model on encoded image sequences. Similar to image generation, video generation has seen both text-based as well as condition-based approaches to control the synthesized output.

\section{Preliminary}
\label{SD-ControlNet}

\noindent\textbf{Latent Diffusion Models~\cite{rombach2021highresolution}} (LDM)~\cite{rombach2021highresolution}, represent those diffusion models uniquely designed to operate within the latent space facilitated by an autoencoder, specifically $\mathcal{D}(\mathcal{E}(\cdot))$. A notable instance of such models is the Stable Diffusion (SD)~\cite{rombach2022high}, which integrates a Vector Quantized-Variational AutoEncoder (VQ-VAE) \cite{van2017neural} and a U-Net structure \cite{ronneberger2015u}. SD employs a CLIP-based transformer architecture as a text encoder~\cite{radford2021learning} to convert text inputs into embeddings, denoted by $c_{\text{text}}$. The training regime of SD entails presenting the model with an image $I$ and a text condition $c_{\text{text}}$. This process involves encoding the image to a latent representation $z_{0}=\mathcal{E}(I)$ and subjecting it to a predefined sequence of $T$ diffusion steps governed by a Gaussian process. This sequence yields a noisy latent representation $z_{T}$, which approximates a standard normal distribution $\mathcal{N}(0,1)$. SD's learning objective is iteratively denoising $z_{T}$ back into the latent representation $z_{0}$, formulated as follows:
\begin{equation}\resizebox{0.9\hsize}{!}{$
    \mathcal{L} = \mathbb{E}_{\mathcal{E}(I), c_{\text{text}},\epsilon\sim\mathcal{N}(0,1),t}\left[\lVert \epsilon-\epsilon_{\theta}(z_t, t, c_{\text{text}}) \rVert_{2}^{2} \right] $}
\end{equation}
where $\epsilon_{\theta}$ is the UNet with learnable parameters $\theta$ and $t=1,...,T$ denotes the time-step embedding in denoising. These modules employ convolutional layers, specifically Residual Blocks (ResNetBlock), and incorporate both self- and cross-attention mechanisms through Transformer Blocks (TransformerBlock).

\noindent\textbf{ControlNet} is an extension of SD that is able to control the generated image layout of SD without modifying the original SD's parameters. It achieves this by replicating the encoder of SD to learn feature residuals for the latent feature maps in SD. It has been successfully applied to different controlled image generation tasks including pose-conditioned human image generation~\cite{zhang2023adding}.
% It enhances its control capabilities by integrating a replicated, trainable configuration of the downsampling and middle blocks from SD-UNet, which are crucial for capturing and synthesizing details. 
% Additionally, the \textit{zero convolution} in upsampling blocks further enables manipulations of structure information, e.g., human pose. During the denoising process, the feature map output from these \textit{zero convolution} layers is integrated with the skip connections according to the original SD-UNet architecture. 

\section{MagicPose}\label{sec:method}
Given a image $I_{R}$ with a person in it, the objective of MagicPose to re-pose the person in the given image to the target pose $\{P, F\}$, where $P$ is the human pose skeleton and $F$ is the facial landmarks. Such a pipeline can be decomposed into two sub-tasks: (1) keeping and transferring the appearance of the human individual and background from reference image and (2) controlling generated images with the pose and expression defined by $\{P, F\}$. To ensure the generazability of the model, MagicPose is designed to inherit the structures and parameters as much as possible from pre-trained stable diffusion models. To this end, we propose an attention-based appearance controller by replicating the structures of the original UNet. An additional ControlNet is then trained jointly to control the pose and expression of the person. We train MagicPose on human video datasets where image pairs of the same person but different poses are available. Then during testing, the reference $I_{R}$ and poses $\{P, F\}$ could come from different sources for pose transfer. The overview of the proposed method (MagicPose) is illustrated in Figure.~\ref{fig:framework}. We first presents our preliminary experiments in terms of appearance control in Sec.~\ref{preliminary_experiments}, which motivates us to propose the Appearance Control Module as elaborated in Sec.~\ref{referenceonly}. Then, Sec.~\ref{pose} presents the fine-tuning of the Appearance-disentangled Pose Control. 
% Finally, we emphasize the major difference between our proposed Appearance Control Model and ControlNet~\cite{zhang2023adding} in Sec.~\ref{comp}.

% Finally, we add an optional motion module, introduced by AnimateDiff~\cite{guo2023animatediff}, for MagicPose to generate sequential output, which will be discussed in Sec.~\ref{mm}.

\subsection{Exploration of Appearance Control Mechanism}\label{preliminary_experiments}
\begin{figure}
\centering
\vspace{-7pt}
 \includegraphics[width=\linewidth]{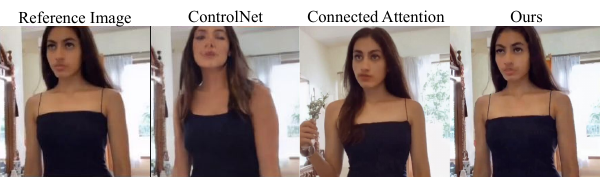}
  \vspace{-25pt}
   \caption{
Identity and appearance control ability comparison between different architectural designs.
   }
   \vspace{-15pt}
    \label{fig:arch_design}
\end{figure}

We first evaluated vanilla ControlNet for appearance control. As shown in Figure~\ref{fig:arch_design}, we found that ControlNet is not able to maintain the appearance when generating human images of different poses, making it unsuitable for the re-targeting task.
On the other side, recent studies~\cite{cao2023masactrl,lin2023consistent123,Zhang} have found that self-attention layers in the diffusion models is highly relevant to the appearance of the generated images. Inspired by them, we conduct an experiment on self-attention for zero-shot appearance control, where the reference image and the noisy image are both forwarded through the diffusion UNet with their self-attention layers connected. A critical observation is that is such an architecture can naturally lead to an appearance resemblance between the two images, even without any fine-tuning (Figure~\ref{fig:arch_design} connected attention). One plausible explanation is that self-attention layers in the UNet plays an important role to transmit the appearance information spatially and hence it could serve as a deformation module to generate similar images with different geometric structures. From another perspective, such an forward process mimics the generation of two image as a single one, and thus, their appearance tend to be similar. However, the problem with such a zero-shot approach is that the generation results are not stable.

\subsection{Appearance Control Pretraining}\label{referenceonly}
Given the above observations, we introduce our Appearance Control Model, which inherits the structure and capability of the zero-shot attention-based control but further extends its stability by introducing task-specific parameters. In particular, it is designed as an auxiliary UNet branch to provide layer-by-layer attention guidance. As shown in Figure.~\ref{fig:framework}, our Appearance Control Model consists of another trainable copy of the original SD-UNet, which connects to the Appearance Control Model by sharing the key and value through the Multi-Source Self Attention Module. 

Formally, the calculation of self-attention in TransformerBlocks of SD-UNet can be written as: 
\begin{equation}
\label{eq:sd_self_attn}
\resizebox{0.5\hsize}{!}{$
   Self\_Attn = softmax(\frac{Q\cdot K^T}{\sqrt{d}})\cdot V $}
\end{equation}
where $Q, K, V$ are query, key, and value. $d$ denotes the dimension of the key and query. In our Multi-Source Self Attention Module, we concatenate the key-value pairs from the Appearance Control Model with SD-UNet together as new key-value pairs and calculate the attention similar to Eq.~\ref{eq:sd_self_attn}:
\begin{equation}\label{eq:our_attn}
\resizebox{0.75\hsize}{!}{$
   Our\_Attn = softmax(\frac{Q_{1}\cdot (K_{1}\oplus K_{2})^T}{\sqrt{d}})\cdot (V_{1}\oplus V_{2})$}
\end{equation}
where $Q_1, K_1, V_1$ are query, key, and value from self-attention layers in the TransformerBlocks of SD-UNet and $K_2, V_2$ are from the Appearance Control Model. $\oplus$ refers to vector concatenation. In essence, the only modification for the SD-UNet is to change the calculation of self-attention from Eq.~\ref{eq:sd_self_attn} to Eq.~\ref{eq:our_attn}.

In order to maintain the generalizability of the SD, in the first training stage (\textbf{Appearance Control Pre-training}), we fix the original UNet and only train the Appearance Control module. The pose ControlNet is not included in this stage. The objective of Appearance Control Pretraining is:
\begin{equation}
\label{eq:acp_obj}
\resizebox{0.9\hsize}{!}{
   $ \mathcal{L} = \mathbb{E}_{\mathcal{E}(I), A_{\theta}(I_{R}), \epsilon\sim\mathcal{N}(0,1),t}\left[\lVert \epsilon-\epsilon_{\theta}(z_t, t, A_{\theta}(I_{R})) \rVert_{2}^{2} \right]$}
\end{equation}
where $A_{\theta}$ is the Appearance Control Model taking reference image $I_{R}$ as input. $\epsilon_{\theta}$ is the SD-UNet, which takes the noisy latent $z_t$, denoising step $t$ and $Our\_Attn$ as inputs.

\subsection{Appearance-disentangled Pose Control}\label{pose}
To control the pose in the generated images, a naive solution directly integrates the pre-trained OpenPose ControlNet model~\cite{zhang2023adding} with our pre-trained Appearance Control Model without fine-tuning. However, our experiments indicate that such a combination struggles with appearance-independent pose control, leading to severe errors between the generated poses and the input poses. To address the issue, we reuse our pre-trained Appearance Control module to disentangle the pose ControlNet from appearance information. In particular, assuming the Appearance Controller already provides a complete guidance for the generated image's appearance, we fine-tune the Pose ControlNet jointly with our Appearance Control Model. As such, Pose ControlNet exclusively modulates the pose attributes of the human, while the Appearance Control Model focuses on appearance control.
% (optional) distill pose ControlNet with image input...
Specifically, we fine-tune MagicPose with an objective similar to latent diffusion training~\citep{rombach2022high}:
\begin{equation}
\resizebox{0.9\hsize}{!}{
   $ \mathcal{L} = \mathbb{E}_{\mathcal{E}(I), A_{\theta}(I_{R}), P_{\theta}(I_{C}),\epsilon\sim\mathcal{N}(0,1),t}\left[\lVert \epsilon-\epsilon_{\theta}(z_t, t, A_{\theta}(I_{R}), P_{\theta}(I_{C})) \rVert_{2}^{2} \right]$}
\end{equation}
where ${P_{\theta}}$ is the Pose ControlNet taking poses $I_{C}$ as inputs.

\begin{figure*}[t!]\vspace{-5pt}
\centering
 \includegraphics[width=\linewidth]{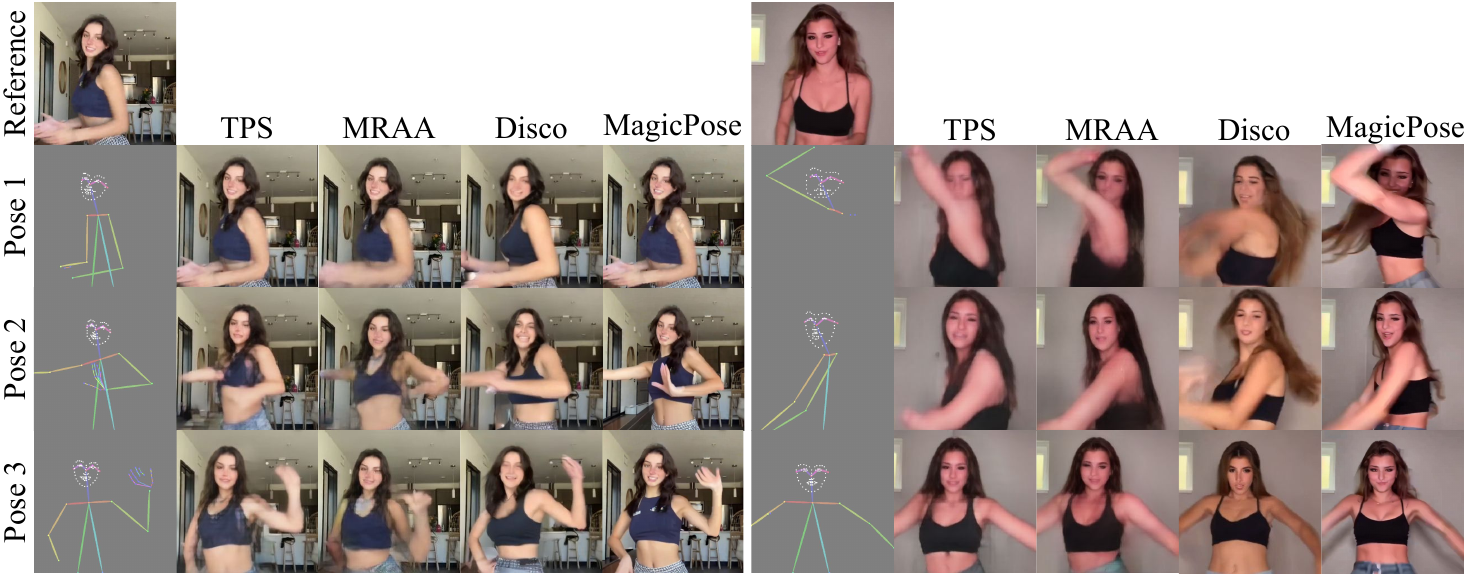}
  \vspace{-25pt}
   \caption{
 Qualitative comparison of human poses and facial expressions retargeting between TPS~\cite{zhao2022thin}, MRAA~\cite{siarohin2019first}, Disco~\cite{wang2023disco} and MagicPose. Previous methods suffer from inconsistent facial expressions and human pose identity. 
   }
   \vspace{-12pt}
    \label{fig:comparison}
\end{figure*}

%$A_{\theta}$ and ${P_{\theta}}$ are the trainable network modules. Specifically, $\epsilon_{\theta}$ contains the U-Net architecture composed of the ResBlocks and TransformerBlocks, which accepts the noisy latent $z_t$ as input. $A_{\theta}$ and ${P_{\theta}}$ represent the Appearance Control Model and Pose ControlNet branches for reference image $I_{R}$ and condition map $I_{C}$ of pose skeleton $P$ and facial landmark $F$.

% \subsection{Motion Module}\label{mm}
% Appearance Control Model and Apperance-disentangled Pose ControlNet together already achieve accurate image-to-image motion transfer, but we can further integrate an optional motion module into the primary SD-UNet architecture to improve the temporal consistency. We initially employed the widely-used AnimateDiff~\cite{guo2023animatediff}, which provides an assortment of motion modules tailored to the stable diffusion model v1.5., but we found that AnimateDiff faces limitations in achieving seamless transition across frames, particularly with more complex movement patterns present in human dance, as opposed to more subdued video content. To solve this issue, we fine-tuned the AnimateDiff motion modules until satisfactory temporal coherence was observed during the evaluation.

\section{Experiments}\label{sec:exps}

\begin{table*}[t!]
\centering
% \captionsetup{font=footnotesize,labelfont=footnotesize,skip=1pt}
\caption{Quantitative comparisons of MagicPose with the recent SOTA methods DreamPose~\cite{karras2023dreampose} and Disco~\cite{wang2023disco}. 
% ``CFG'' and ``HAP'' denote classifier-free guidance and human attribute pretraining, respectively. 
$\downarrow$ indicates that the lower the better, and vice versa. Methods with $*$ directly use the target image as the input, including more information compared to the OpenPose~\cite{8765346,simon2017hand,cao2017realtime,wei2016cpm}. $^\dagger$ represents that Disco~\cite{wang2023disco} is pre-trained on other datasets~\cite{fu2022stylegan,ge2019deepfashion2,schuhmann2021laion,lin2014microsoft} more than our proposed MagicPose, which uses only 335 video sequences in the TikTok~\cite{Jafarian_2021_CVPR_TikTok} dataset for pretraning and fine-tuning. \textbf{Face-Cos} represents the cosine similarity of the extracted feature by AdaFace~\cite{kim2022adaface} of face area between generation and ground truth image.
}
\label{tab:quant_comp_tiktok}
{\begin{tabular}{lccccccc}
\toprule
\multirow{2}[2]{*}{Method}  & \multicolumn{6}{c}{\textbf{Image}} & \multicolumn{1}{c}{\textbf{Video}} \\ \cmidrule(lr){2-7} \cmidrule(lr){8-8}
 & \textbf{FID}\ $\downarrow$  & \textbf{SSIM}\ $\uparrow$ & \textbf{PSNR}\ $\uparrow$ & \textbf{LPIPS}\ $\downarrow$ & \textbf{L1} $\downarrow$\ & \textbf{Face-Cos}\ $\uparrow$ & \textbf{FID-VID}\  $\downarrow$ \\
\midrule
FOMM$^*$~\citep{siarohin2019first}  &85.03  &0.648  &29.01  & 0.335  &3.61E-04  &0.190 &90.09 \\
MRAA$^*$~\citep{siarohin2021motion}  &54.47  &0.672  & 29.39  &0.296  &3.21E-04  & 0.337 &66.36 \\
TPS$^*$~\citep{zhao2022thin}  &53.78  &0.673  & 29.18  &0.299  &3.23E-04  &0.280 &72.55  \\
DreamPose~\citep{karras2023dreampose} &72.62    &0.511  &28.11  &0.442 &6.88E-04 & 0.085&78.77  \\ 
DisCo~\cite{wang2023disco} & 50.68    &0.648  &28.81  &0.309  &4.27E-04  & - &69.68  \\
DisCo$^\dagger$~\cite{wang2023disco} &30.75    &0.668  &29.03  &\textbf{0.292}  &3.78E-04  & 0.166 &59.90  \\

\midrule

MagicPose  &\textbf{25.50}   &\textbf{0.752 } &\textbf{29.53}  & \textbf{0.292}  &\textbf{0.81E-04} & \textbf{0.426}
 &\textbf{46.30}  \\
\bottomrule
\end{tabular}}
\vspace{-5pt}

\end{table*}

\subsection{Datasets}\label{dataset}
\par \noindent \textbf{TikTok}~\cite{Jafarian_2021_CVPR_TikTok} dataset consists of 350 single-person dance videos (with video length of 10-15 seconds). Most of these videos contain the face and \textit{\textbf{upper-body}} of a human. For each video, we extract frames at 30fps and run OpenPose~\cite{8765346,simon2017hand,cao2017realtime,wei2016cpm} on each frame to infer the human pose skeleton, facial landmarks, and hand poses. 335 videos are sampled as the training split. We follow~\cite{wang2023disco} and use their 10 TikTok-style videos depicting different people from the web as the testing split. 

\par \noindent \textbf{Everybody Dance Now}~\cite{chan2019everybody} consists of \textbf{full-body} videos of five subjects. Experiments on this dataset aim to test our method's generalization ability to in-the-wild, full-body motions.

\par \noindent \textbf{Self-collected Out-of-Domain Images} come from online resources. We use them to test our method's generalization ability to in-the-wild appearance.

\subsection{Implementation Details}\label{implement}
We first pre-train the appearance control model on 8 NVIDIA A100 GPUs with batch size 192 for 10k steps with image size $512 \times 512$ and learning rate $0.0001$. We then jointly fine-tune the appearance and pose control model on 8 NVIDIA A100 GPUs with batch size 16 for 20K steps. The Stable-Diffusion UNet weight is frozen during all experiments. During training, we randomly sampled the two frames of the video as the reference and target. Both reference and target images are randomly cropped at the same position along the height dimension with the aspect ratio of 1 before resizing to $512 \times 512$. For evaluation, we apply center cropping instead of random cropping.
We initialize the U-Net model with the pre-trained weights of Stable-Diffusion Image Variations~\cite{sd-image-variations}. The Appearance Control Model branch is initialized with the same weight as the U-Net model. After Appearance Control pre-training, we initialize the U-Net and Appearance Control Model branch with the previous pre-trained weights and initialize the Pose ControlNet branch with the weight from~\cite{zhang2023adding}, for joint fine-tuning.
After these steps, an optional motion module can be further fine-tuned.

% We finally freeze the weights of all aforementioned parts in our pipeline and fine-tune the motion module with pre-trained weights from AnimateDiff~\cite{guo2023animatediff} for 30k steps with a batch size of 8. Each batch contains 16 frames of a video sequence as the target output.

\subsection{Qualitative and Quantitative Comparison}\label{comparison}

We conduct a comprehensive evaluation of TikTok~\cite{Jafarian_2021_CVPR_TikTok} in comparison to established motion transfer methodologies, including FOMM~\citep{siarohin2019first}, MRAA~\citep{siarohin2021motion}, and TPS~\citep{zhao2022thin}, as well as recent advancements in the field such as
% DreamPose~\citep{karras2023dreampose} and 
Disco~\cite{wang2023disco}. Disco~\cite{wang2023disco} leverages a CLIP encoder to integrate appearance information from the reference image into the Transformer Blocks of the Stable-Diffusion UNet and Pose ControlNet while retaining OpenPose~\cite{8765346,simon2017hand,cao2017realtime,wei2016cpm} as the pose condition. Though OpenPose has the limitation of incomplete detection of the human skeleton (More details in supplementary), we follow previous work and adopt OpenPose as the pose detector. For image quality evaluation, we adhere to the methodology outlined in Disco~\cite{wang2023disco} and report metrics such as frame-wise FID~\citep{heusel2017gans}, SSIM~\citep{wang2004image}, LPIPS~\citep{zhang2018unreasonable}, PSNR~\citep{hore2010image}, and L1. 
In addition to these established metrics, we introduce a novel image-wise metric called \textbf{Face-Cos}, which stands for Face Cosine Similarity. This metric is designed to gauge the model's capability to preserve the identity information of the reference image input. To compute this metric, we first align and crop the facial region in both the generated image and the ground truth. Subsequently, we calculate the cosine similarity between the extracted feature by AdaFace~\cite{kim2022adaface}, frame by frame of the same subject in the test set, and report the average value as the \textbf{Face-Cos} score. The experimental results, in Table~\ref{tab:quant_comp_tiktok}, unequivocally establish MagicPose as a front-runner, showcasing substantial improvements across all metrics compared to alternative approaches. Notably, MagicPose attains a Face-Cos score of $\sim 0.426$, representing a substantial \textbf{+0.260} enhancement over Disco. This performance shows MagicPose's robust capacity to preserve identity information. The substantial performance improvement over the state-of-the-art methods demonstrates the superiority of the MagicPose framework.
We qualitatively compare MagicPose to previous methods~\cite{zhao2022thin,siarohin2019first,wang2023disco} in Figure~\ref{fig:comparison}. 
TPS~\cite{zhao2022thin}, MRAA~\cite{siarohin2019first}, and Disco~\cite{wang2023disco} suffer from inconsistent facial expressions and human appearances. 
% We provide more examples of motion and facial expression transfer in Figure~\ref{fig:motion}. 
Please check the supplementary materials to see more examples of real-human poses and facial expressions re-targeting. 

% DreamPose is tailored for the fashion domain, incorporating the idea of replacing the CLIP text feature in the diffusion model with an image embedding through a dual CLIP-VAE encoder and adapter module. Moreover, it employs a sequence of denseposes~\cite {guler2018densepose} instead of the conventional OpenPose~\cite{8765346,simon2017hand,cao2017realtime,wei2016cpm} for pose conditioning. 
% follows a similar paradigm as DreamPose,

%Conversely, for video evaluation, we aggregate every consecutive set of 16 frames to generate a sample and report the FID-VID~\citep{balaji2019conditional}.
\begin{table}[t!]\vspace{-3pt}
\centering
% \captionsetup{font=footnotesize,labelfont=footnotesize,skip=1pt}
\caption{User study of MagicPose. We collect the number of votes from 100 participants for eight subjects in the test set. The participants found that MagicPose preserves the best identity and appearance information in pose and facial expression retargeting.
}
\label{tab:user_study}
{\begin{tabular}{lcccccccccccc}
\toprule
Method & {\textbf{Average}}\\
\midrule
MRAA~\cite{siarohin2019first}  &4\% \\
FOMO~\cite{siarohin2021motion} &3\% \\
TPS~\cite{zhao2022thin}  & 4\% \\
Disco~\cite{wang2023disco}   &16\%  \\
MagicPose  &\textbf{73\%}\\
\bottomrule
\end{tabular}}
\vspace{-20pt}
\end{table}

\begin{figure*}[t!] \vspace{-3pt}
\centering
 \includegraphics[width=\linewidth]{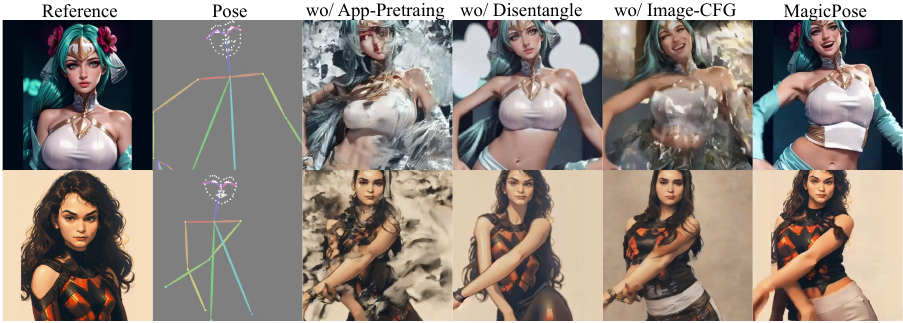}
 \vspace{-22pt}
   \caption{
Ablation Analysis of MagicPose. The proposed Appearance Control Pretraining and Appearance-disentangled Pose Control provide better identity control and generation quality effectively.
   }
    \label{fig:ablation}
    \vspace{-10pt}
\end{figure*}

\begin{table*}[t]
\centering
% \captionsetup{font=footnotesize,labelfont=footnotesize,skip=1pt}
\caption{Ablation Analysis of MagicPose with different training and inference settings. \textbf{App-Pretrain} stands for Appearance Control Pretraining through Multi-Source Attention Module and \textbf{Disentangle} denotes Appearance-disentangled Pose Control. \textbf{Image-CFG} denotes classifier free guidance. \textbf{Data Aug} indicates the model is trained with data augmentation of random masking of facial landmarks and hand poses. 
}
\label{tab:ablation_image}
\setlength\extrarowheight{1pt}
\scalebox{0.8}
{\begin{tabular}{cccc ccccccc}
\toprule
  \multirow{2}[2]{*}{App-Pretrain}  &  \multirow{2}[2]{*}{Disentangle} &\multirow{2}[2]{*}{Image CFG.} & \multirow{2}[2]{*}{Data Aug.} & \multicolumn{6}{c}{\textbf{Image}} & \multicolumn{1}{c}{\textbf{Video}} \\ \cmidrule(lr){5-10} \cmidrule(lr){11-11}&&&
 & \textbf{FID}\ $\downarrow$  & \textbf{SSIM}\ $\uparrow$ & \textbf{PSNR}\ $\uparrow$ & \textbf{LPIPS}\ $\downarrow$ & \textbf{L1} $\downarrow$\ & \textbf{Face-Cos}\ $\uparrow$ & \textbf{FID-VID}\  $\downarrow$ \\
\midrule
\xmark & \xmark& \cmark & \cmark &  288.64 &0.291 & 27.85 & 0.695 & 2.69E-04 &0.038 &382.10 \\
\cmark & \xmark& \cmark & \cmark &  35.76  &0.727 & 28.79 & 0.321  & 0.97E-04& 0.397 & 65.72\\
\cmark & \cmark& \xmark & \cmark &   61.33  & 0.659 & 28.31 &  0.357 &1.29E-04 &0.272 & 98.96 \\
\cmark & \cmark& \cmark & \xmark &28.71   &0.751 &29.14  &0.296  &0.86E-04 & 0.417
 &47.26 \\
\cmark & \cmark& \cmark & \cmark & \textbf{25.50}   &\textbf{0.752 } &\textbf{29.53}  & \textbf{0.292}  &\textbf{0.81E-04} & \textbf{0.426}
 &\textbf{46.30}  \\
\bottomrule
\end{tabular}}
\vspace{-10pt}
\end{table*}

\noindent \textbf{User Study} We provide a  user study for comparison between MagicPose and previous works~\cite{siarohin2019first,siarohin2021motion,zhao2022thin,wang2023disco}. We collect reference images, openpose conditions, and pose retargeting results from prior works and MagicPose of 8 subjects in the test set. For each subject, we visualize different human poses and facial expressions and ask 100 users to choose \textbf{only one} method, which preserves the best identity and appearance information for each subject. We present the averaged vote result in Table.~\ref{tab:user_study}. Visualization examples and detailed user studies can be found in supplementary material.

\subsection{Ablation Analysis}\label{ablation}
In this section, a comprehensive ablation analysis of MagicPose on the TikTok~\cite{Jafarian_2021_CVPR_TikTok} dataset is presented. The impact of various training and inference configurations within MagicPose is systematically analyzed in Table~\ref{tab:ablation_image}. We examine the proposed Appearance Control Model and its Multi-Source Self-Attention Module, specifically assessing their contributions when omitted. The absence of Appearance Control Pretraining and Appearance-disentangled Pose Control reveals the significance of these components, which can be observed in Figure.~\ref{fig:ablation} as well. Notably, the introduction of Appearance Control Pretraining markedly enhances generation quality, evidenced by a substantial increase of \textbf{+944.73\%} in Face-Cos and \textbf{+149.82\%} in SSIM. Additionally, the implementation of Appearance-disentangled Pose Control demonstrates its efficacy, yielding improvements of \textbf{+7.30\%} in Face-Cos and \textbf{+3.43\%}  in SSIM. 
Furthermore, we highlight the necessity of incorporating the data augmentation technique of randomly masking facial landmarks and hand poses during training. This is particularly crucial due to the occasional limitations of OpenPose~\cite{8765346,simon2017hand,cao2017realtime,wei2016cpm} in providing complete and accurate detection of hand pose skeletons and facial landmarks, which can result in artifacts in generated images. Therefore, to enhance the robustness of MagicPose against incomplete human pose estimations by OpenPose~\cite{8765346,simon2017hand,cao2017realtime,wei2016cpm}, this data augmentation strategy is proposed and leads to incremental improvements in Face-Cos and SSIM by \textbf{+2.20\%} and \textbf{+0.13\%}, respectively. 
Moreover, the application of classifier-free guidance (Image-CFG) in the training process, as discussed in prior work~\cite{wang2023disco,Ho2022ClassifierFreeDG,Lin2023CommonDN,Balaji2022eDiffITD,dao2022flashattention} on diffusion models, further augments the quality of generation. The implementation of Image-CFG enhances Face-Cos by \textbf{+56.62\%} and SSIM by \textbf{+14.11\%}, underscoring its value in the image generation context.

\begin{table*}[t!]\vspace{-5pt}
\centering
% \captionsetup{font=footnotesize,labelfont=footnotesize,skip=1pt}
\caption{Quantitative evaluation of generalization ability of MagicPose. MagicPose$\dagger$ denotes the pipeline is directly evaluated on test set of Everybody Dance Now~\cite{chan2019everybody} after being trained on TikTok~\cite{Jafarian_2021_CVPR_TikTok}, and MagicPose$\ddagger$ represents the pipeline is further fine-tuned on Everybody Dance Now~\cite{chan2019everybody} train set and evaluated on test set. 
}
\label{tab:edn_generalization}
\scalebox{0.85}
{\begin{tabular}{lcccccccccccc}
\toprule
\multirow{2}[2]{*}{Method}  & \multicolumn{2}{c}{\textbf{Subject1}} & \multicolumn{2}{c}{\textbf{Subject2}}& \multicolumn{2}{c}{\textbf{Subject3}}& \multicolumn{2}{c}{\textbf{Subject4}}& \multicolumn{2}{c}{\textbf{Subject5}} & \multicolumn{2}{c}{\textbf{Average}} \\ \cmidrule(lr){2-3} \cmidrule(lr){4-5} \cmidrule(lr){6-7} \cmidrule(lr){8-9} \cmidrule(lr){10-11} \cmidrule(lr){12-13}
 & \textbf{FID}\ $\downarrow$  & \textbf{PSNR}\ $\uparrow$  & \textbf{FID}\ $\downarrow$  & \textbf{PSNR}\ $\uparrow$ & \textbf{FID}\ $\downarrow$  & \textbf{PSNR}\ $\uparrow$ & \textbf{FID}\ $\downarrow$  & \textbf{PSNR}\ $\uparrow$ & \textbf{FID}\ $\downarrow$  & \textbf{PSNR}\ $\uparrow$ & \textbf{FID}\ $\downarrow$  & \textbf{PSNR}\ $\uparrow$\\
\midrule
MagicPose$\dagger$  & 22.59  & \textbf{30.67} & \textbf{22.21}  & \textbf{30.13 }  &35.43 &\textbf{29.35}& \textbf{31.72}  & 29.53  & 31.24 & 28.48 & 28.64 & 29.63 \\
MagicPose$\ddagger$  &\textbf{22.50}   & \textbf{30.67}& 22.61  & 28.40 & \textbf{27.38} & 29.10
 &  36.73 &\textbf{33.95}  & \textbf{21.99} &\textbf{30.94 } & \textbf{26.24} & \textbf{30.61}\\
\bottomrule
\end{tabular}}
\vspace{-10pt}
\end{table*}

% Image style
\begin{figure}[t]
\centering
 \includegraphics[width=\linewidth]{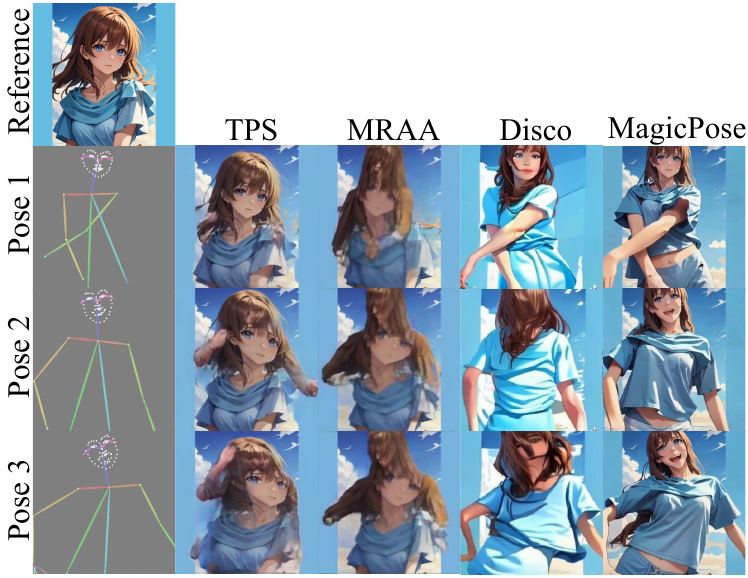}
    \vspace{-20pt}
   \caption{
Comparison of zero-shot pose and facial expression
retargeting on out-of-domain image.
   }
    \label{fig:gen_comp}
\vspace{-10pt}
\end{figure}

% Ethnicity
\begin{figure}
\centering
 \includegraphics[width=\linewidth]{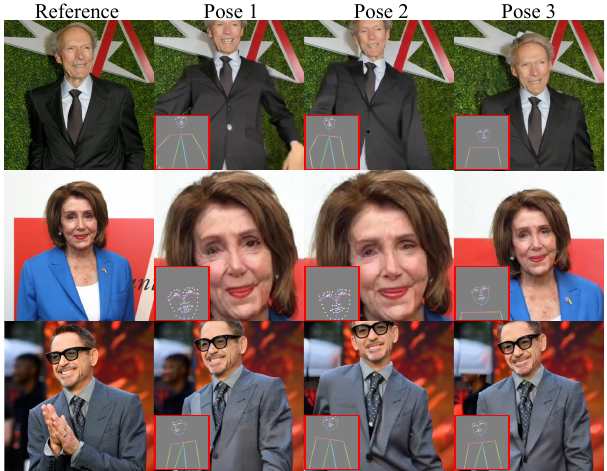}
 \vspace{-20pt}
   \caption{
        Visualization of zero-shot pose and facial expression retargeting on in-the-wild real-human with different ethnicity and age from training data (Tiktok).
    }
    \label{fig:ethical}
    \vspace{-10pt}
\end{figure}

% Everybody Dance Now
\begin{figure}[t!]
\centering
 \includegraphics[width=\linewidth]{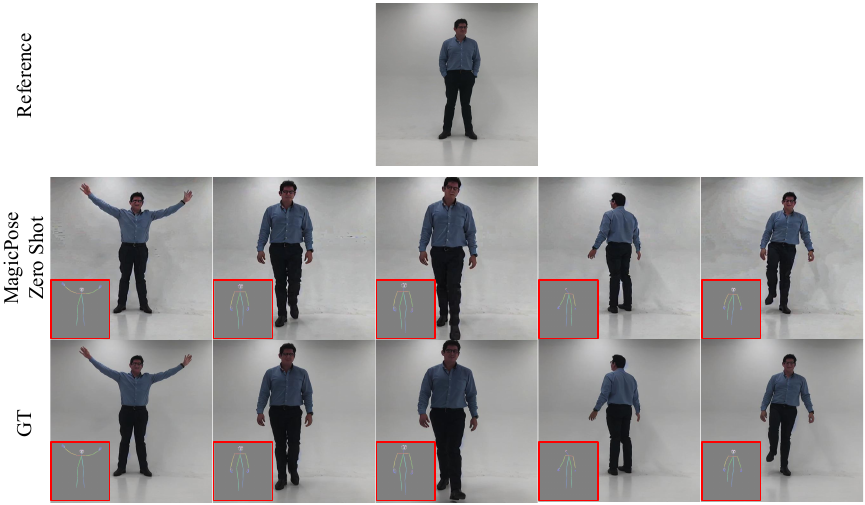}
 \vspace{-15pt}
   \caption{
Visualization of zero-shot Human Motion and Facial Expression Transfer on EverybodyDancNow Dataset~\cite{chan2019everybody}. MagicPose can generalize perfectly to in-the-wild human motions.
   }
    \label{fig:edn}
    \vspace{-10pt}
\end{figure}

\subsection{Generalization Ability}
% \noindent \textbf{Generalization Ability}
It is also worth highlighting that MagicPose can generalize to out-of-domain reference images of unseen styles and poses with surprisingly good appearance controllability, even without any further fine-tuning on the target domain.
% , while previous works~\cite{siarohin2019first,siarohin2021motion,zhao2022thin,wang2023disco} cannot provide satisfying results.
Figure.~\ref{fig:gen_comp} compares the zero-shot results of applying TPS~\cite{zhao2022thin}, MRAA~\cite{siarohin2019first}, Disco~\cite{wang2023disco} and MagicPose to out-of-domain images, whose visual style is distinct from corresponding training data of the real-human upper-body images. For real-human reference images, we observe that most of the human subjects from TikTok~\cite{Jafarian_2021_CVPR_TikTok} dataset and the self-collected test set of Disco~\cite{wang2023disco} are young women. So we test our method on more in-the-wild real-human examples, e.g. elder people, in Figure~\ref{fig:ethical}. We also evaluate the in-the-wild motions generalization ability of MagicPose on Everybody Dance Now~\cite{chan2019everybody}, which is a full-body dataset, in contrast to the upper-body images used in the TikTok dataset. We directly apply MagicPose to such full-body reference images and visualize the qualitative results in Figure.~\ref{fig:edn} and provide a quantitative evaluation in Table.~\ref{tab:edn_generalization}.  Experiments show that MagicPose generalizes surprisingly well to full body images even though it has never been trained on such data. Furthermore, better quality of generation can be achieved after fine-tuning on specific datasets as well. More visualizations of zero-shot Animation and results on Everybody Dance Now~\cite{chan2019everybody} can be found in the supplementary materials.

\section{Conclusion}\label{sec:conclusion}
In this work, we propose MagicPose, a novel approach in the realm of realistic human poses and facial expressions retargeting. By seamlessly incorporating motion and facial expression transfer and enabling the generation of consistent in-the-wild animations without any further fine-tuning, MagicPose shows a significant advancement over prior methods. Notably, our approach demonstrates a superior capacity to generalize over diverse human identities and complex motion sequences. Moreover, MagicPose boasts a practical implementation as a plug-in module or extension compatible with existing models such as Stable Diffusion~\cite{rombach2022high}. This combination of innovation, efficiency, and adaptability establishes MagicPose as a promising tool in the field of poses and facial expressions retargeting.
\section{Acknowledgements}
Soleymani's work was sponsored by the Army Research Office and was accomplished under Cooperative Agreement Number W911NF-20-2-0053. The views and conclusions contained in this document are those of the authors and should not be interpreted as representing the official policies, either expressed or implied, of the Army Research Office or the U.S. Government. The U.S. Government is authorized to reproduce and distribute reprints for Government purposes notwithstanding any copyright notation herein.

% \clearpage
% \newpage
\section*{Impact Statement}
The proposed MagicPose for retargeting human poses and facial expressions has a wide range of applications. It can significantly improve communication in digital environments, enabling individuals to express themselves more effectively through avatars or digital characters, thereby enhancing interactions in virtual meetings, online classrooms, and social networking platforms. Additionally, MagicPose has the potential to revolutionize entertainment and media production, allowing for the creation of more lifelike and expressive characters in movies, video games, and animations, consequently fostering more immersive storytelling experiences and increased audience engagement. The experiment demonstrates our model can generalize across different real-human ethnicities and ages, and even to out-of-domain images, e.g., cartoon-style images, painting-style images, and AI-generated images.

Potential Negative Social Impact: The method can potentially be used in a malicious way, e.g., making fake animated videos of people, which could be used in fraud. To avoid the potential misuse of such technologies, It is essential to employ several solutions like digital watermarking and detection algorithms, enact and enforce stringent legal measures, enhance public awareness and education on media literacy, and establish ethical guidelines within the tech industry. This involves collaboration among tech companies, governments, educators, and the public to create a safer digital environment and mitigate the risks of fraudulent AI-generated content

% \clearpage
% \newpage
{
    \small
    
    \bibliographystyle{icml2024}
    \bibliography{main}
}
\clearpage
\setcounter{page}{1}
\appendix

\section{Detailed User Study}

\begin{table*}[htp]
\centering
\caption{The user study with 100 participants. We collect the number of votes for eight video subjects from test set by five methods and report the percentage. Our MagicPose preserves the best identity information in pose and facial expression retargeting on all subjects.
}
\vspace{5pt}
\label{tab:user_study_rebuttal}
\setlength\extrarowheight{1pt}
\scalebox{0.85}
{\begin{tabular}{lcccccccccccc}
\toprule
Method  & {\textbf{Subject1}} & {\textbf{Subject2}}& {\textbf{Subject3}}& {\textbf{Subject4}}&{\textbf{Subject5}} & {\textbf{Subject6}} & {\textbf{Subject7}}& {\textbf{Subject8}}& {\textbf{Average}}\\
\midrule
MRAA~\cite{siarohin2019first}  & 8\%  & 6\% & 0\%  & 5\%    &2\% &2\% &8\%   & 4\%  &4\%  \\
FOMO~\cite{siarohin2021motion}  & 3\%  & 1\% & 3\% & 1\%  &1\%   &0\% & 5\%  & 8\%  &3\% \\
TPS~\cite{zhao2022thin}  & 4\%    & 16\% & 0\%  & 4\%    &2\% &3\%   & 4\%    & 2\% & 4\%  \\
Disco~\cite{wang2023disco}  & 12\%& 3\%  & 9\% & 18\%  & 5\%  &20\% &33\% & 27\% & 16\%  \\
MagicPose  & \textbf{73\%}  & \textbf{74\%} & \textbf{88\%}  & \textbf{72\%}  &\textbf{90\%} &\textbf{75\%} & \textbf{50\%}  &\textbf{59\%} &\textbf{73\%}\\
\bottomrule
\end{tabular}}
\end{table*}

In this section, we provide a comprehensive user study for qualitative comparison between MagicPose and previous works~\cite{siarohin2019first,siarohin2021motion,zhao2022thin,wang2023disco}. As we mentioned in the experiment, we collect reference images, openpose conditions, and pose retargeting results from prior works and MagicPose of 8 subjects in the test set. For each subject, we visualize different human poses and facial expressions. Some examples are shown in Figure.~\ref{fig:user_study_1} and Figure.~\ref{fig:user_study_2}. The methods are anonymized as A, B, C, D, E, and the order of the generated image from the corresponding method is randomized in each subject comparison. We ask 100 users to choose \textbf{only one} method which preserves the best identity and appearance information for each subject. We present the full result in Table.~\ref{tab:user_study_rebuttal}.

\textbf{Participants}
We use \textbf{Prolific}, an online platform designed to connect researchers with participants for academic studies and market research for all our user studies. The participants are English-speaking random users verified by this platform without prior knowledge of computer vision.

\textbf{Criteria for Judgment}
Since our paper focuses on the motion retargeting task, the criteria for evaluation are 1) The appearance (Face, Clothes on the body, Background) of the generation should strictly match the given reference image input. 2) The motions and facial expressions of the generation should strictly match the given pose condition map input. As mentioned in Section A of our Supplementary Material, We ask the participants to choose \textbf{the only one} method which preserves the best identity information for each video subject. 
In order to compare in a professional manner, all the methods are anonymized as A, B, C, D, E, and the order of the generated image from the corresponding method is randomized in each subject comparison, e.g. For comparison of video subject 1, A,B,C,D,E correspond to MRAA, FOMO, TPS, Disco, MagicPose. For comparison of video subject 2, A,B,C,D,E correspond to Disco, MRAA, MagicPose, FOMO, TPS.

\textbf{Results}
In order to make the conclusion from our user study statistically significant, we recruited 100 participants from Prolific. The result is presented in Table.~\ref{tab:user_study_rebuttal} and we can conclude that MagicPose preserves the identity the best compared to prior works.

\textbf{Statistical Analysis}
We perform a chi-square test on our results. For each vote by an participant, we label the chosen method as 1 and the rest methods as 0. Results of such a test are shown in Table.~\ref{tab:user_study_rebuttal}, we compare five methods (A, B, C, D, E) on eight video subjects with the following steps:
\begin{enumerate}
    \item State the \textit{Null Hypothesis}: There is no association between the video subjects and the choice of method. The distribution of votes for each method is the same across all groups, meaning any observed differences are due to chance.

    \item Compute chi-square statistic and p-value: Chi-square statistic: \(116.02\). p-value: \(1.11 \times 10^{-12}\). Degrees of freedom: \(28\).

    \item Conclusion and Discussion: Given the extremely small p-value (much less than 0.05), we can reject the Null Hypothesis. This indicates that there is a statistically significant association between the video subjects and the choice of method. In simpler terms, the differences in vote distribution for the methods across the eight groups are unlikely to have occurred by chance, pointing towards a significant preference pattern among the groups. We conclude that the participants indeed prefer our proposed MagicPose more than other methods.
\end{enumerate}

\section{Additional Visulizations}
\subsection{TikTok}
\begin{figure*}[t!]
\centering
 \includegraphics[width=0.95\linewidth]{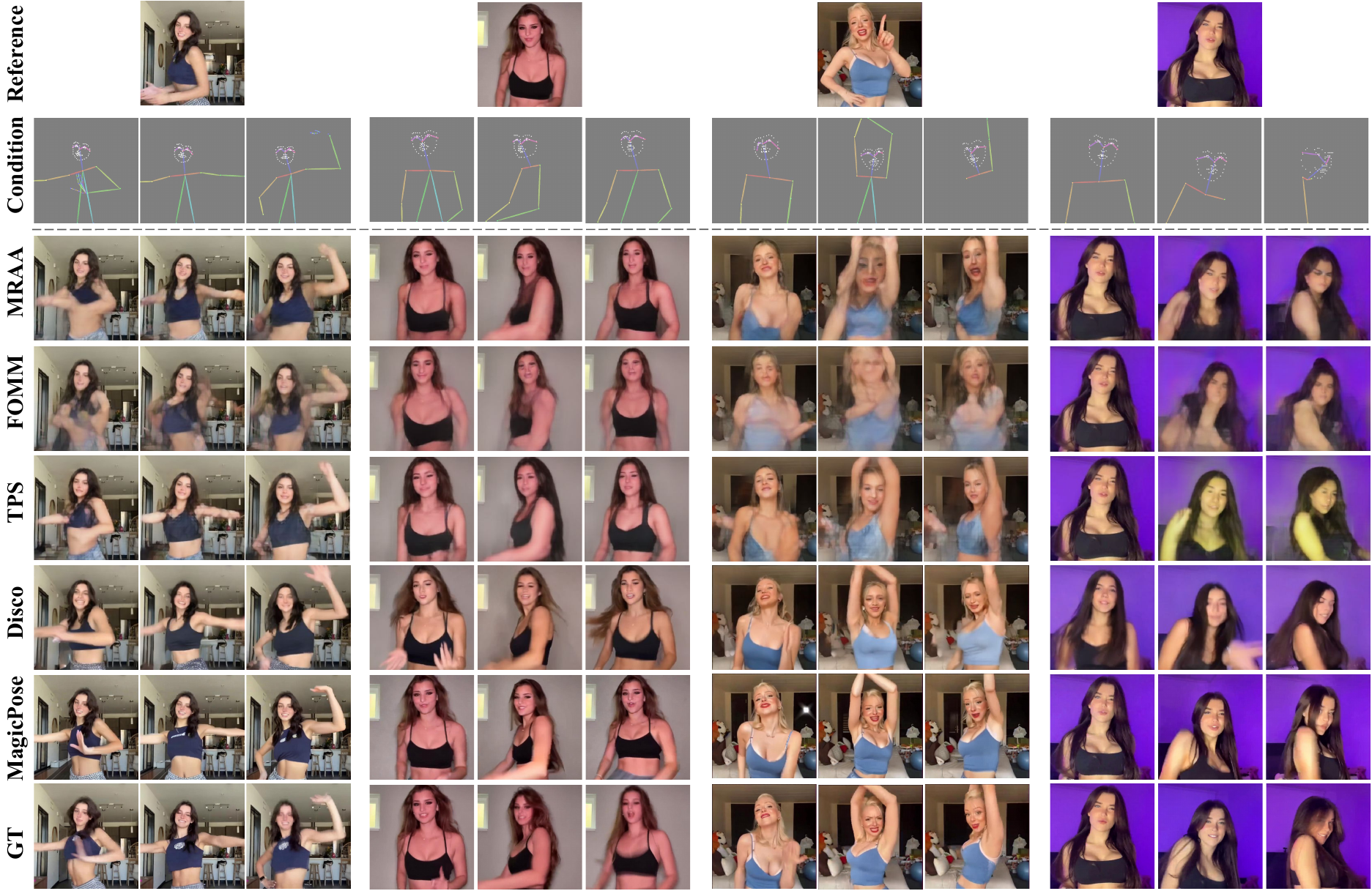}
   \caption{
Visualization of human pose and facial expression retargeting (Subjects 1-4 in the user study): MagicPose preserves identity and appearance information from the reference image the best.
   }
    \label{fig:user_study_1}
\end{figure*}

\begin{figure*}[t!]
\centering
 \includegraphics[width=0.95\linewidth]{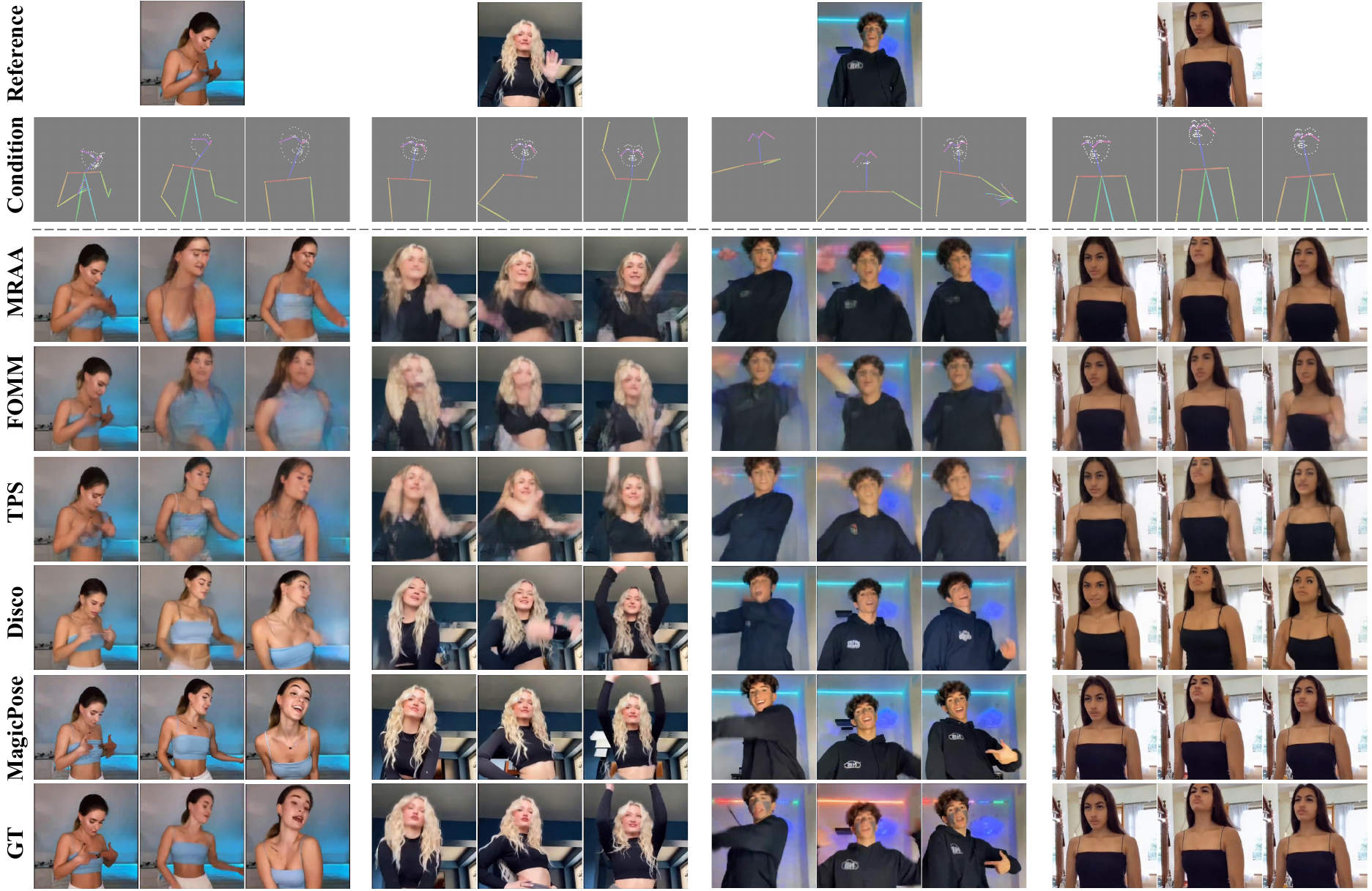}
   \caption{
Visualization of human pose and facial expression retargeting (Subjects 5-8 in the user study): MagicPose preserves identity and appearance information from the reference image the best.
   }
    \label{fig:user_study_2}
\end{figure*}

\begin{figure*}[t!]
\centering
 \includegraphics[width=0.95\linewidth]{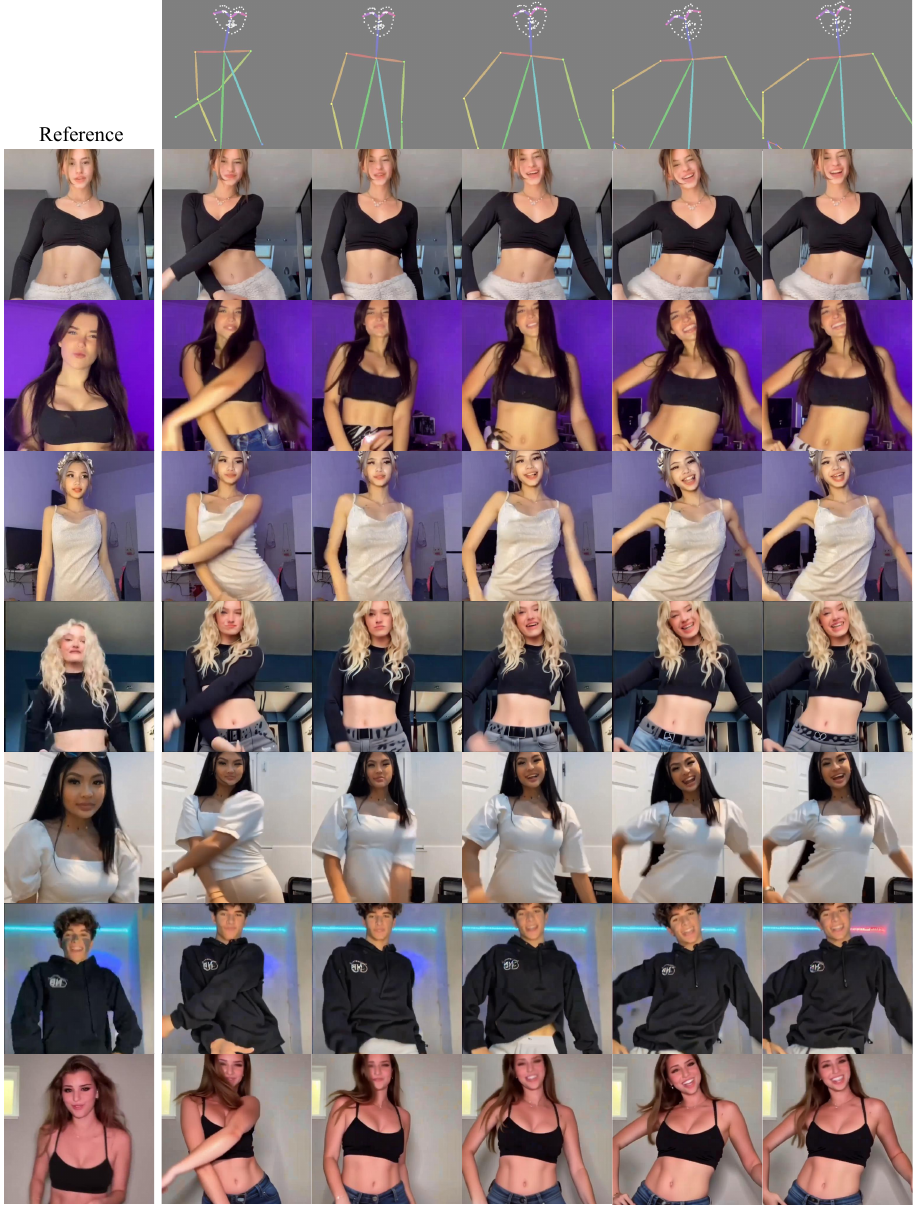}
   \caption{
Visualization of Human Motion and Facial Expression Transfer on TikTok~\cite{Jafarian_2021_CVPR_TikTok}.
   }
    \label{fig:tiktok_supp1}
\end{figure*}

\begin{figure*}[t!]
\centering
 \includegraphics[width=\linewidth]{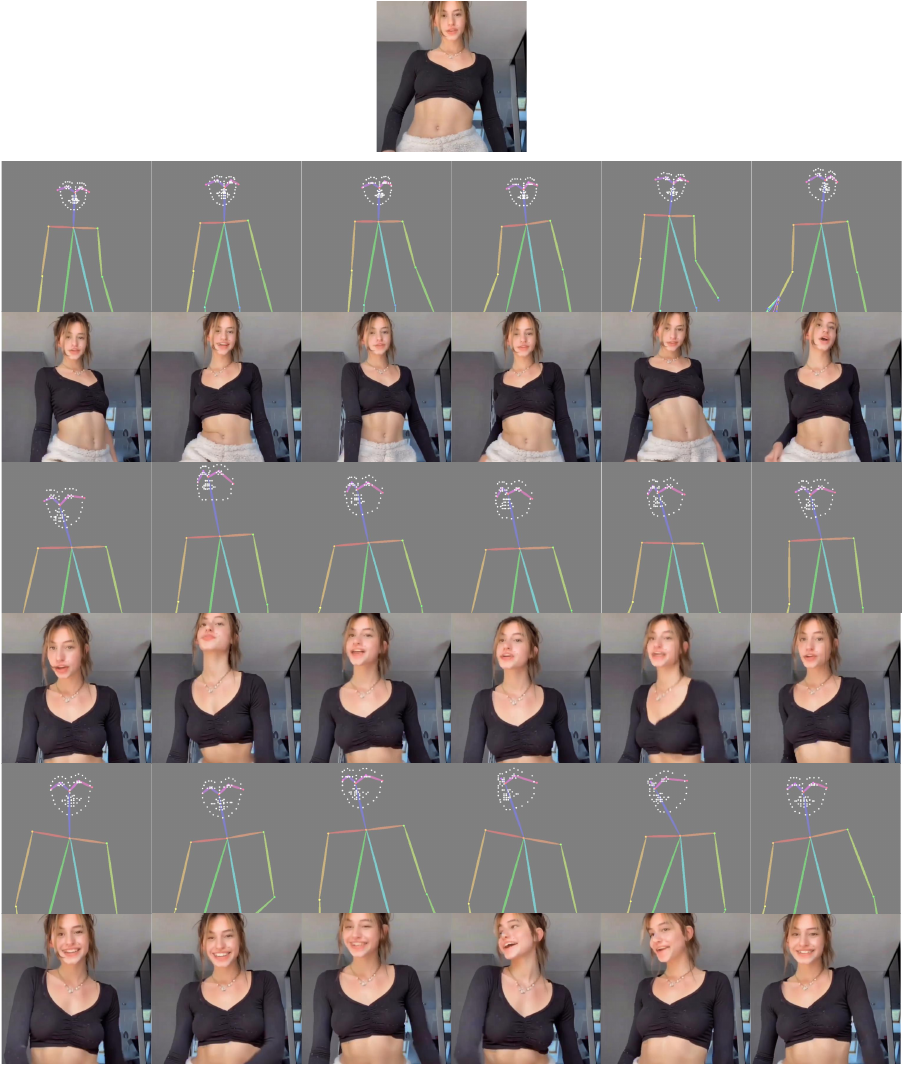}
   \caption{
Visualization of Human Motion and Facial Expression Transfer on TikTok~\cite{Jafarian_2021_CVPR_TikTok}. MagicPose is able to generate vivid and realistic motion and expressions under the condition of diverse pose skeleton and face landmark input, while accurately maintaining identity information from the reference image input.
   }
    \label{fig:tiktok_supp2}
\end{figure*}

\begin{figure*}[t!]
\centering
 \includegraphics[width=\linewidth]{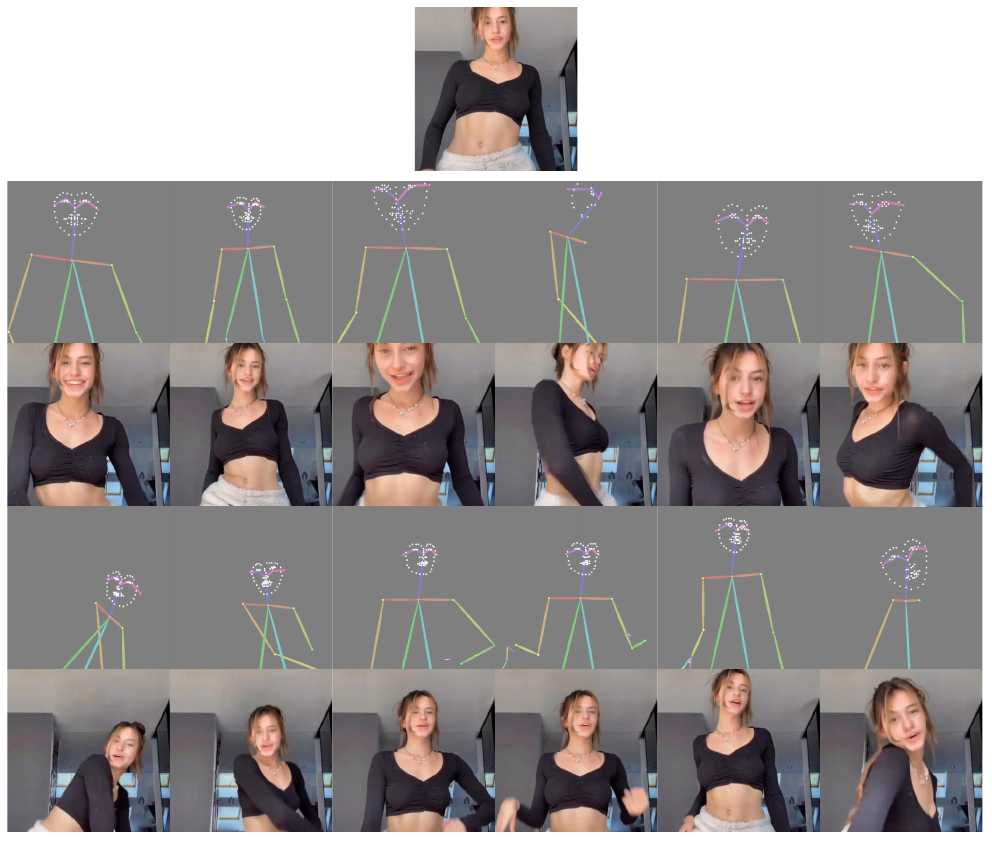}
  \vspace{-25pt}
   \caption{
Visualization of Human Motion and Facial Expression Transfer on TikTok~\cite{Jafarian_2021_CVPR_TikTok}. 
   }
   %\vspace{-10pt}
    \label{fig:motion}
\end{figure*}

We provide more visualizations on the test set of the experiments on TikTok~\cite{Jafarian_2021_CVPR_TikTok} in Figure.~\ref{fig:user_study_1}, Figure.~\ref{fig:user_study_2}, Figure.~\ref{fig:tiktok_supp1}, Figure.~\ref{fig:tiktok_supp2}, and Figure.~\ref{fig:motion}.

\begin{figure}[ht]
\centering
 \includegraphics[width=\linewidth]{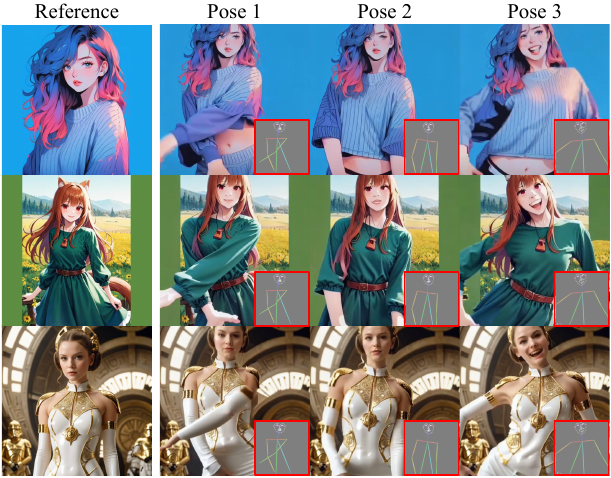}
\vspace{-10pt}
   \caption{
Visualization of generalization to unseen image styles that are different from our training set (Tiktok).
   }
    \label{fig:animie}

\end{figure}

% We also provide cross-subject pose retargeting results of real-human images, zero-shot cartoon images from our model and let the user to grade the quality. (From 0-10)

% \begin{table*}[t!]
% \centering
% % \captionsetup{font=footnotesize,labelfont=footnotesize,skip=1pt}
% \caption{User study of MagicPose. We collect the number of votes for eight subjects in the test set and report the percentage. The participants found that MagicPose preserves the best identity and appearance information in pose and facial expression retargeting.
% }
% \label{tab:user_study_full}
% \setlength\extrarowheight{1pt}
% \scalebox{0.9}
% {\begin{tabular}{lcccccccccccc}
% \toprule
% Method  & {\textbf{Subject1}} & {\textbf{Subject2}}& {\textbf{Subject3}}& {\textbf{Subject4}}&{\textbf{Subject5}} & {\textbf{Subject6}} & {\textbf{Subject7}}& {\textbf{Subject8}}& {\textbf{Average}}\\
% \midrule
% MRAA~\cite{siarohin2019first}  & 10\%  & 10\% & 0\%  & 3\%    &0\% &0\% &3\%   & 3\%  &3\%  \\
% FOMO~\cite{siarohin2021motion}  & 5\%  & 0\% & 8\% & 0\%  &3\%   &0\% & 5\%  & 10\%  &4\% \\
% TPS~\cite{zhao2022thin}  & 3\%    & 18\% & 0\%  & 3\%    &0\% &3\%   & 3\%    & 0\% & 3\%  \\
% Disco~\cite{wang2023disco}  & 13\%& 3\%  & 10\% & 18\%  & 10\%  &28\% &43\% & 28\% & 19\%  \\
% MagicPose  & \textbf{70\%}  & \textbf{70\%} & \textbf{83\%}  & \textbf{78\%}  &\textbf{88\%} &\textbf{70\%} & \textbf{48\%}  &\textbf{60\%} &\textbf{71\%}\\

% \bottomrule
% \end{tabular}}

% %%%%\vspace{-5pt}
% \end{table*}

\subsection{EverybodyDanceNow}
\begin{figure*}[t!]
\centering
 \includegraphics[width=\linewidth]{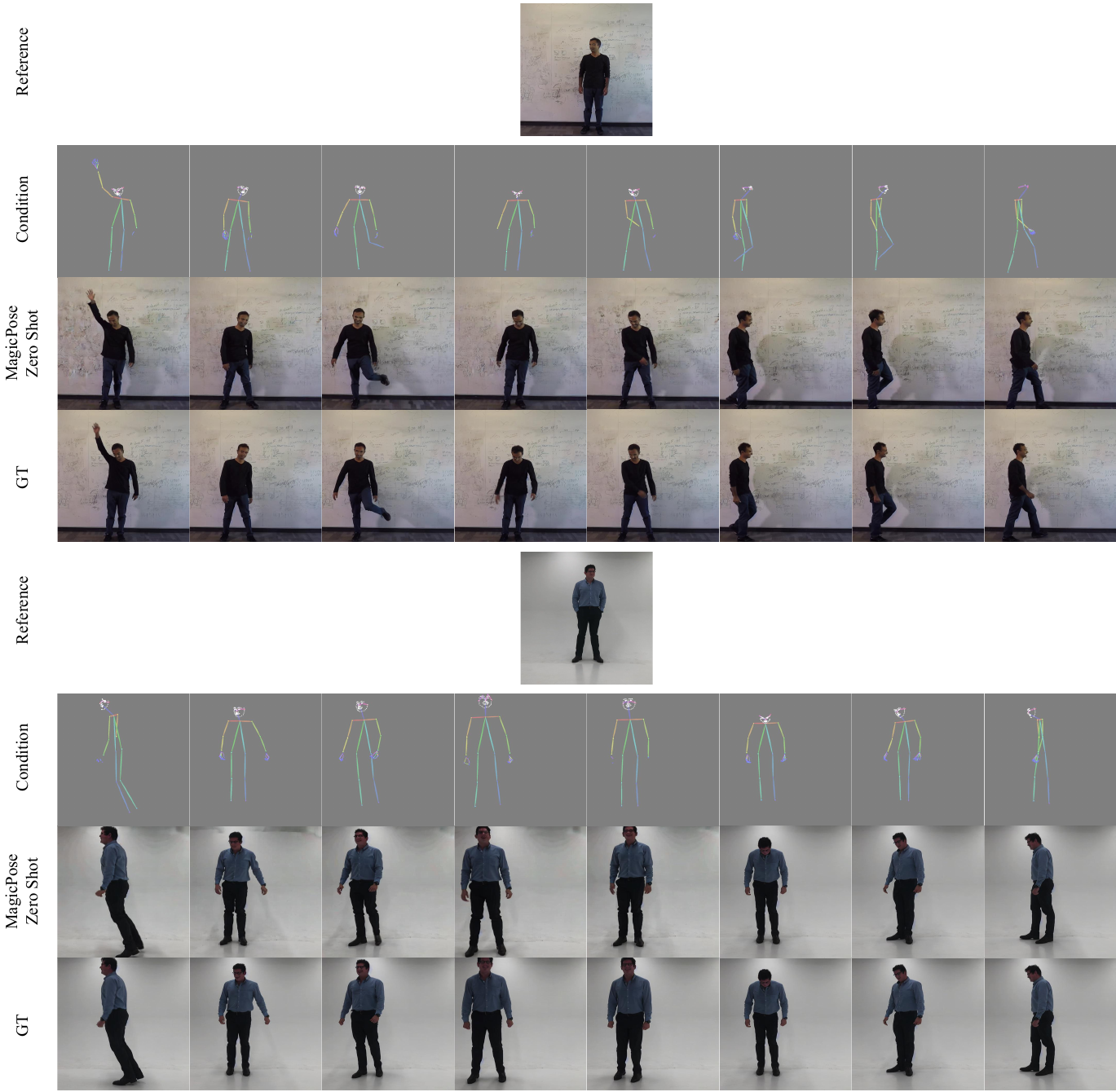}
   \caption{
Visualization of Zero-Shot Human Motion and Facial Expression Transfer on Everybody Dance Now Dataset~\cite{chan2019everybody}.
   }
    \label{fig:edn_supp1}
\end{figure*}

\begin{figure*}[t!]
\centering
 \includegraphics[width=\linewidth]{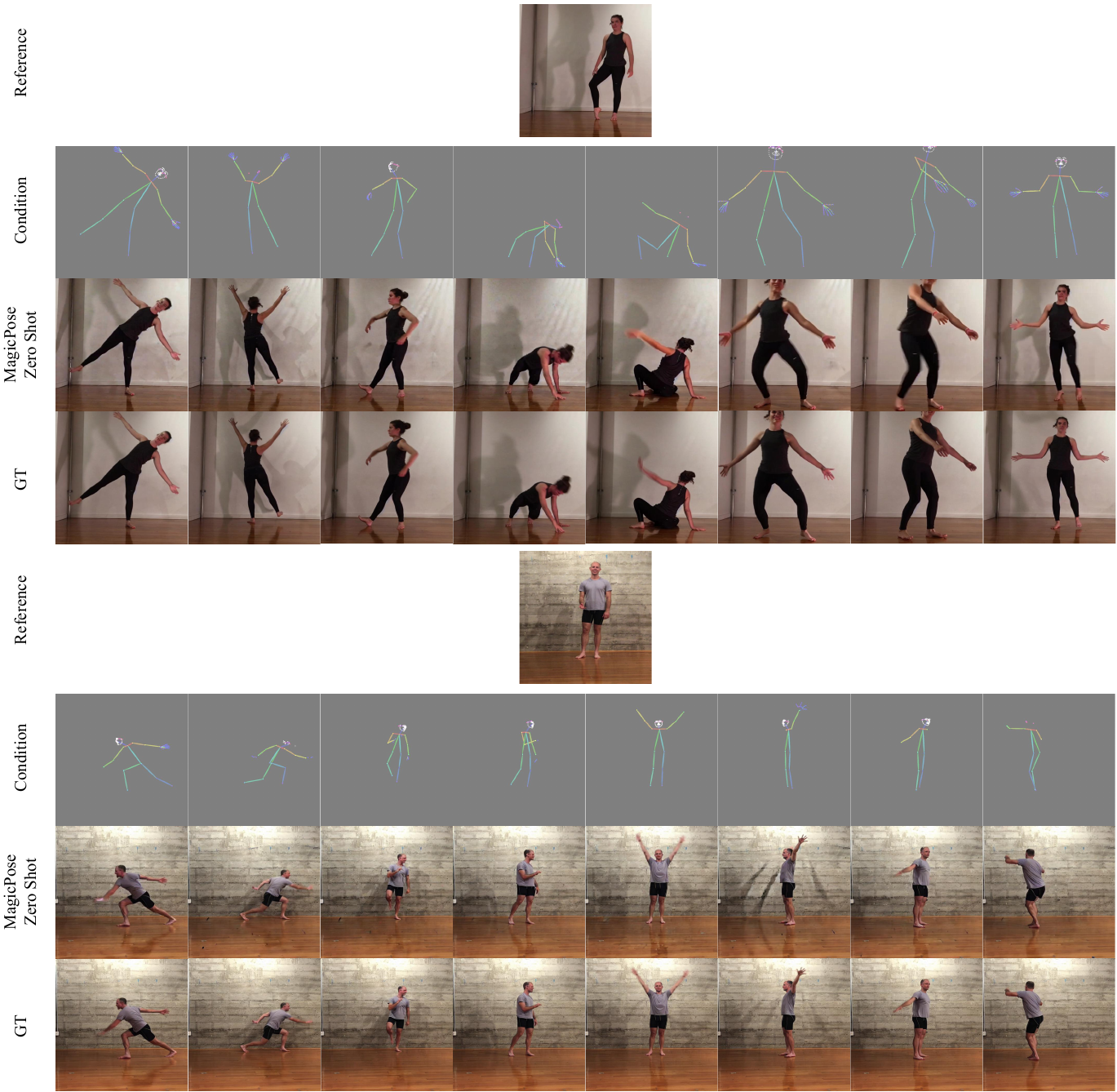}
   \caption{
Visualization of Zero-Shot Human Motion and Facial Expression Transfer on Everybody Dance Now Dataset~\cite{chan2019everybody}.
   }
    \label{fig:edn_supp2}
\end{figure*}

We provide more visualizations of zero-shot generation on Everybody Dance Now dataset~\cite{chan2019everybody} in Figure.~\ref{fig:edn_supp1} and Figure.~\ref{fig:edn_supp2}.

\subsection{Zero-Shot Animation}
\subsubsection{Out-of-domain Images}
\begin{figure*}[t!]
\centering
 \includegraphics[width=0.85\linewidth]{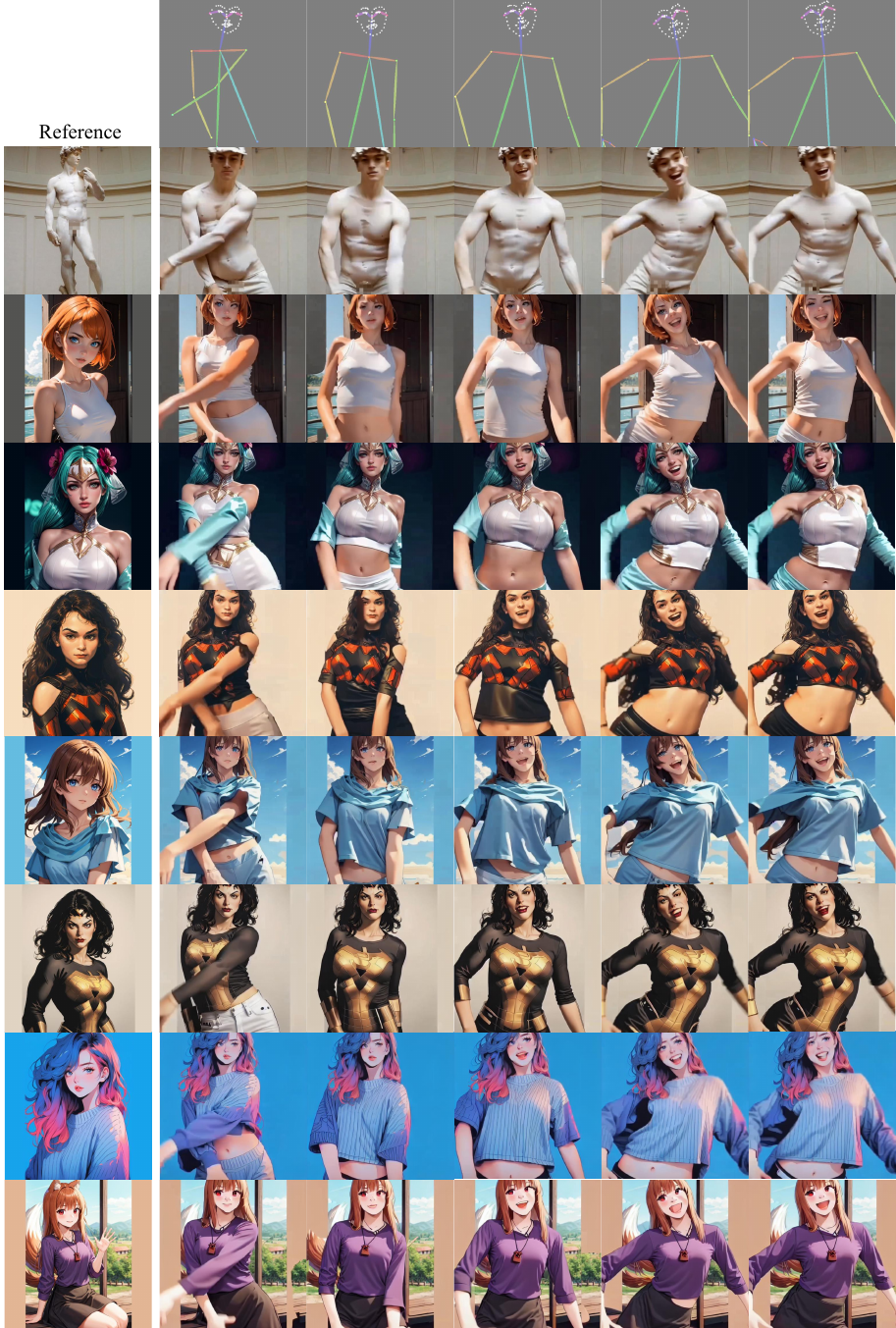}
   \caption{
Visualization of Zero-Shot Animation. MagicPose can provide a precise generation with identity information from out-of-domain images even without any further fine-tuning after being trained on real-human dance videos.  
   }
    \label{fig:cartoon1}
\end{figure*}

\begin{figure*}[t!]
\centering
 \includegraphics[width=0.95\linewidth]{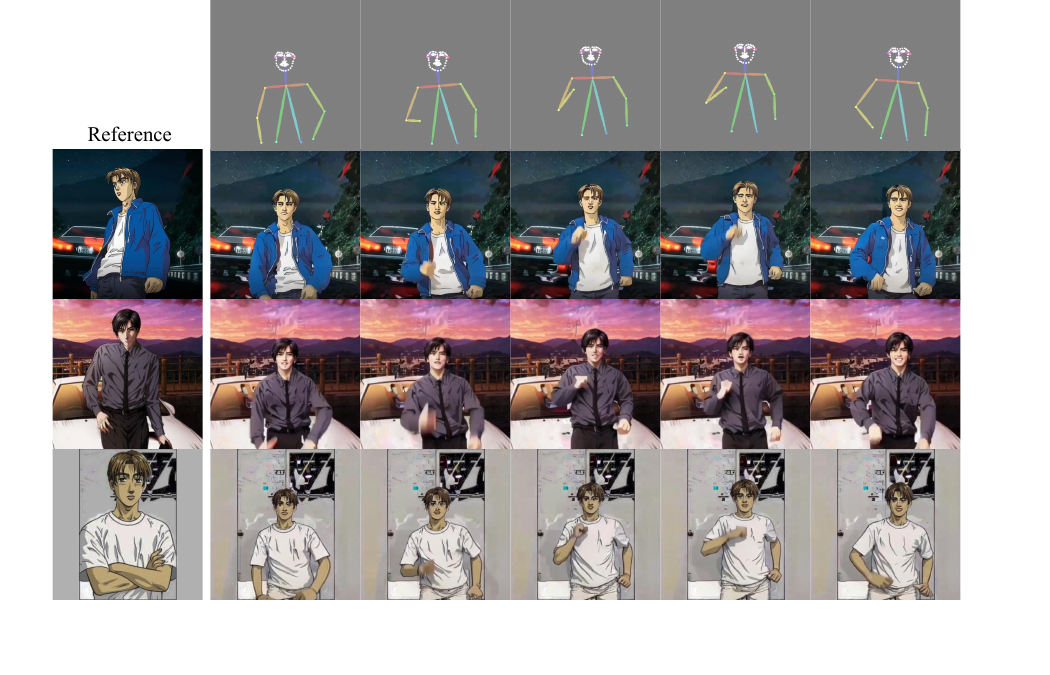}
   \caption{
Visualization of Zero-Shot 2D Cartoon Generation.   
   }
    \label{fig:cartoon2}
\end{figure*}

\begin{figure*}[t!]
\centering
 \includegraphics[width=\linewidth]{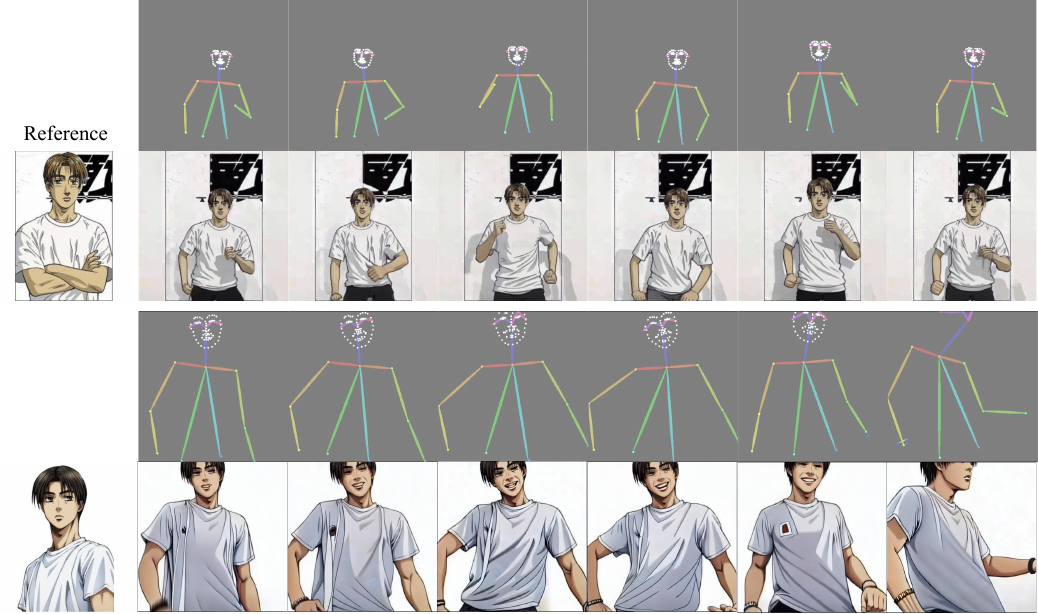}
   \caption{
Visualization of Zero-Shot 2D Cartoon Generation.  
   }
    \label{fig:cartoon3}
\end{figure*}

We provide more visualizations of zero-shot generation of out-of-domain images in  Figure.~\ref{fig:animie}, Figure.~\ref{fig:cartoon1}, Figure.~\ref{fig:cartoon2}, and Figure.~\ref{fig:cartoon3}.

% Image style


\subsubsection{Combine with T2I Model}
\begin{figure*}[t!]
\centering
 \includegraphics[width=0.7\linewidth]{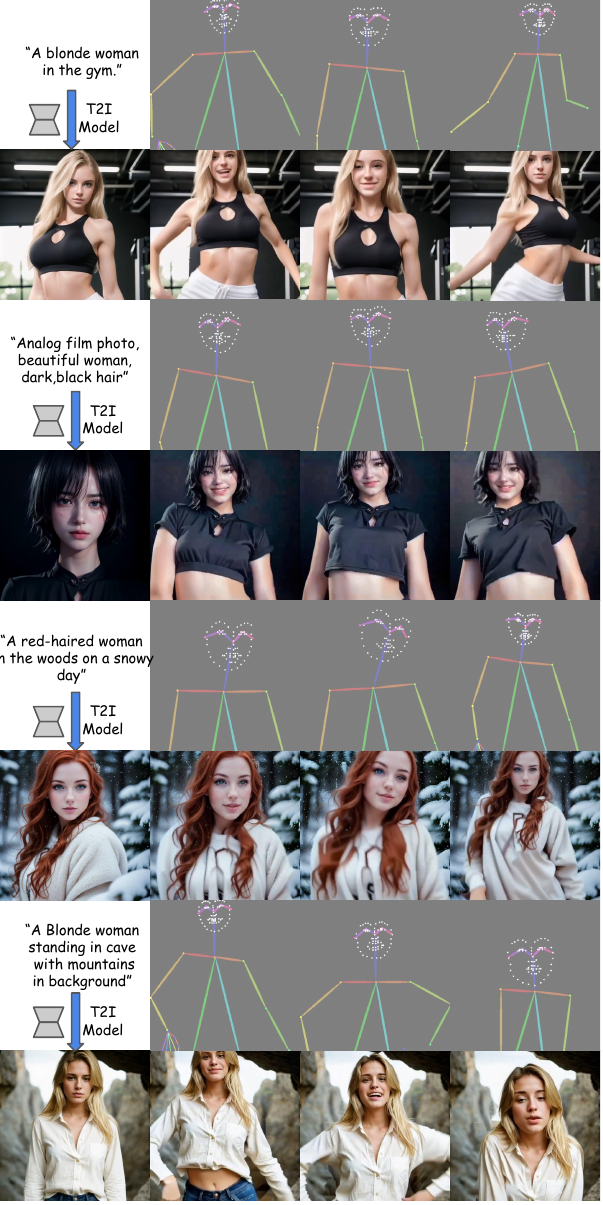}
   \caption{
Usage of combining with T2I model. MagicPose can provide a precise generation with identity information from T2I-generated images even without further fine-tuning after training on real-human dance videos.  
   }
    \label{fig:T2I}
\end{figure*}

A potential application of our proposed model is that it can be combined with the existing Text to Image (T2I) generation model~\cite{zhang2023adding,rombach2022high} and used to edit the generation result. We visualized some samples in Figure.~\ref{fig:T2I}.

% \subsubsection{Visualization on other ethical  human examples}
% We observe that most of the human subjects from TikTok~\cite{Jafarian_2021_CVPR_TikTok} dataset and the self-collected test set of Disco~\cite{wang2023disco} are young women. So in order to test the generalization ability of our model, we have already provided additional results of zero-shot animation of cartoon-style images or AI-generated images by the T2I model in previous sections. Here, we consider more in-the-wild real-human examples, e.g. elder people, in Figure~\ref{fig:ethical}.

% \begin{figure*}[]
% \centering
%  \includegraphics[width=0.95\linewidth]{Figures/Supp/Figure11_Crop.pdf}
%    \caption{
% Visualization of zero-shot pose and facial expression retargeting on in-the-wild real-human elder people. MagicPose can provide a precise generation with identity information from these subjects without any further fine-tuning after being trained on TikTok~\cite{Jafarian_2021_CVPR_TikTok}.}
%     \label{fig:ethical}
% \end{figure*}

\section{Sequence Generation with Motion Module}

As mentioned in our main paper, the Appearance Control Model and Apperance-disentangled Pose ControlNet together already achieve accurate image-to-image motion transfer, but we can further integrate an optional motion module into the primary SD-UNet architecture to improve the temporal consistency. We initially employed the widely-used AnimateDiff~\cite{guo2023animatediff}, which provides an assortment of motion modules tailored to the stable diffusion model v1.5., but we found that AnimateDiff faces limitations in achieving seamless transition across frames, particularly with more complex movement patterns present in human dance, as opposed to more subdued video content. To solve this issue, we fine-tuned the AnimateDiff motion modules until satisfactory temporal coherence was observed during the evaluation. We freeze the weights of all parts in our Appearance Control Model and Apperance-disentangled Pose ControlNet, and fine-tune the motion module with pre-trained weights from AnimateDiff~\cite{guo2023animatediff} for 30k steps with a batch size of 8. Each batch contains 16 frames of a video sequence as the target output. For more smooth and consistent video generation quality, we also propose a special sampling strategy for DDIM~\cite{song2021denoising} during inference.
Figure.~\ref{fig:tiktok_supp1}, Figure.~\ref{fig:cartoon1}, Figure.~\ref{fig:cartoon2}, and Figure.~\ref{fig:cartoon3} are examples of sequential output from our model.

% \section{Applications and Future Works}

\section{Limitations}

In MagicPose, We follow previous work~\cite{wang2023disco} and adopt  OpenPose~\cite{8765346,simon2017hand,cao2017realtime,wei2016cpm} as the human pose detector, which is crucial for pose control, significantly affecting the generated images' quality and temporal consistency. However, challenges arise in accurately detecting complete pose skeletons and facial landmarks, especially under rapid movement, occlusions, or partial visibility of subjects. As illustrated in the second row of Figure~\ref{fig:comparison}, we can observe that the skeleton and hand pose are partially missing in the detection result, especially in the right half of the row. In future works, a more advanced pose detector can be adopted for better image editing quality. 

\section{Discussion on motivation and future works}
In addition to the suggestion of replacing openpose with a more advanced pose detector, we also would like to discuss future works from our motivation. Our understanding of image generation is that it can be decomposed into two aspects: (1) identity control (appearance of human) and (2) shape/geometry control (pose and motion of human). MagicPose was introduced to maintain the appearance and identity information in generation from reference image input strictly while editing the geometry shape and structural information under the guidance of human pose skeleton. In this paper, we demonstrate the \textbf{identity-preserving} ability of the Appearance Control Model and its Multi-Source Attention Module by human pose and facial expression re-targeting task. The design of this Multi-Source Attention Module can be further extended to other tasks as well, e.g. novel view synthesis of general objects under the shape condition of the camera, shape manipulation of the natural scenes under the geometry condition of depth/segmentation map, and motion transfer of animals under the animal pose condition of skeletons, etc.

\section{Comparison to prior works}
\textbf{Comparison with ControlNet} The proposed Appearance Control Model is novel and different in many ways from ControlNet~\cite{zhang2023adding}. In term of control objective, ControlNet was introduced to control the geometrical shape and structural information in the text-to-image model, while our appearance Control Model aims to provide identity and appearance information for the generated subject regardless of the given text. In term of structure design, ControlNet copies the encoder and middle blocks of SD-UNet, whose output feature maps are added to the decoder of SD-UNet to realize \textbf{pose} control. On the other side, the proposed Appearance Control Model 
replicates a whole UNet model to controls the generation process of pre-trained diffusion model via attention layers, enabling more flexible information interchange among distant pixels. And therefore it is more suited for the task of pose retargeting.\\

\textbf{Comparison with MasaCtrl and Reference Only ControlNet}
Both MasaCtrl~\cite{cao2023masactrl} and Reference Only ControlNet~\cite{Zhang} are inference-only models and require text as appearance guidance input. 
MagicPose is a pipeline that can be fine-tuned on customized data and provide consistent identity-preserving generation without any text prompt.

\textbf{MasaCtrl} also utilizes parallel UNet architecture, however, there are several major differences between MasaCtrl and MagicPose 
\begin{enumerate}
    \item The self-attention key-value pairs from the reference branch in MasaCtrl \textbf{replace} those in SD-UNet, while MagicPose's \textbf{Multi-Source} Self-Attention Module concatenates the key-value pairs from both SD-UNet and appearance control model. 
    \item The replacement of key-value pairs only exists in \textbf{decoder} after specific denoising timestep $S$ and after specific layer index $L$ in MasaCtrl, while MagicPose manipulates all self-attention layers in the encoder, middle block, and the decoder. For both inference and training, the manipulation always exists regardless of timesteps. This ensures our model learns both \textbf{encoding} appearance from the reference image (encoder) and \textbf{generating} identity-preserving results(decoder) from customized data.
    \item Text is required as input to generate an extraction \textbf{mask} for key-value pairs in MasaCtrl, while MagicPose doesn't need any additional text information so that appearance information \textbf{only} comes from the reference image. This further helps our model to strictly preserve the identity and make our approach suitable for the motion retargeting (usually there's no text prompt for this task, since pose map defines the human motion and reference image controls human appearance and background).
\end{enumerate}

\textbf{Reference Only ControlNet} does not have a parallel UNet architecture like the original ControlNet~\cite{zhang2023adding}. Reference Only ControlNet shares the same architecture and weight as SD-UNet and first takes a noisy reference image as the only input. During the denoising process of the reference image, the query key and value from the self-attention layers are saved temporarily in a \textbf{\textit{cache}}. Then the text is fed as input to the SD-UNet again and the denoising process yields image generation output, while the self-attention layers take query key value from the \textbf{\textit{cache}}. In contrast, MagicPose introduces a \textbf{trainable} parallel UNet architecture without text input and the appearance control model implicitly learns how to provide identity control for SD-UNet with Multi-Sourse Self-Attnetion Module during fine-tuning.

\end{document}